\UseRawInputEncoding
\documentclass[10pt,journal,compsoc]{IEEEtran}
\usepackage{color}
\usepackage{booktabs}
\usepackage{subfigure}
\usepackage[justification=centering]{caption}
\usepackage{ifpdf}
\ifCLASSINFOpdf
    \usepackage[pdftex]{graphicx}
    \graphicspath{{../pdf/}{../jpeg/}}
\else
    \usepackage[dvips]{graphicx}
    \graphicspath{{../eps/}}
\fi
\usepackage{picins}
\usepackage{amsmath}
\usepackage{acronym}
\usepackage{algorithmic}
\usepackage{array}
\usepackage{mdwmath}
\usepackage{mdwtab}
\usepackage{eqparbox}
\usepackage{stfloats}
\usepackage{hyperref}
\usepackage{url}
\usepackage{color}

\begin{document}

\title{Adversarial Examples Detection with Enhanced Image Difference Features based on Local Histogram Equalization}

\author{Zhaoxia Yin,~\IEEEmembership{Member,~IEEE, }%
        Shaowei Zhu, %
        Hang Su,~\IEEEmembership{Member,~IEEE, }%
        Jianteng Peng,
        Wanli Lyu 
        and~Bin Luo,~\IEEEmembership{Senior Member,~IEEE,}
\IEEEcompsocitemizethanks{\IEEEcompsocthanksitem  Zhaoxia Yin is with the School of Communication \& Electronic Engineering, East China Normal University,Shanghai,China\\E-mail: zxyin@cee.ecnu.edu.cn\\
\IEEEcompsocthanksitem Hang Su is with the Department of Computer Science and Technology, Tsinghua University, Beijing, 100084 China\protect\\
Hang Su is corresponding author. E-mail: suhangss@mail.tsinghua.edu.cn\\
\IEEEcompsocthanksitem Jianteng Peng is currently a senior computer vision algorithm engineer in OPPO Intellisense and Interaction Research Department.\\Email: pengjianteng@oppo.com \\
\IEEEcompsocthanksitem Shaowei Zhu, Wanli Lyu and Bin Luo are with Anhui University \\E-mail: zhusw520@gmail.com, \{wanly lv, ahu\_lb\}@163.com}\\
\thanks{Manuscript received August, 2022}%
}%

\markboth{SUBMITTED TO IEEE TDSC}%
{}

\IEEEtitleabstractindextext{
\begin{abstract}
Deep Neural Networks (DNNs) have recently made significant progress in many fields. However, studies have shown that DNNs are vulnerable to adversarial examples, where imperceptible perturbations can greatly mislead DNNs even if the full underlying model parameters are not accessible. 
Various defense methods have been proposed, such as feature compression and gradient masking. However, numerous studies have proven that previous methods create detection or defense against certain attacks, which renders the method ineffective in the face of the latest unknown attack methods. 
The invisibility of adversarial perturbations is one of the evaluation indicators for adversarial example attacks, which also means that the difference in the local correlation of high-frequency information in adversarial examples and normal examples can be used as an effective feature to distinguish the two. Therefore, we propose an adversarial example detection framework based on a high-frequency information enhancement strategy, which can effectively extract and amplify the feature differences between adversarial examples and normal examples. Experimental results show that the feature augmentation module can be combined with existing detection models in a modular way under this framework. Improve the detector's performance and reduce the deployment cost without modifying the existing detection model.
\end{abstract}

}

\maketitle

\IEEEdisplaynontitleabstractindextext
\IEEEpeerreviewmaketitle
\ifCLASSOPTIONcompsoc
\IEEEraisesectionheading{\section{Introduction} \label{Sec:Introduction}}

\IEEEPARstart{I}{n} recent years, deep neural networks (DNNs) have made significant progress in many fields, such as image classification \cite{fenlei}, speech recognition \cite{Speech}, autonomous driving \cite{Autonomous}, face recognition \cite{face}, and medical diagnosis \cite{Diagnosis,zhenduan}. However, these DNN-based models can be easily attacked by adversarial examples\cite{szegedy2014intriguing} with small perturbations resulting in wrong outputs, that is well known as adversarial attack.

Existing adversarial examples generation methods can be broadly categorized into three categories: (1) gradient descent-based methods, such as fast gradient sign method (FGSM) \cite{goodfellow2014explaining}, iterative FGSM (I-FGSM) \cite{kurakin2017adversarial} and project gradient descent (PGD) \cite{madry2018towards}; (2) optimization-based generative methods, such as C\&W \cite{carlini2017towards} and DeepFool \cite{moosavi2016deepfool}; (3) universal adversarial perturbation (UAP) \cite{moosavi2017universal}. Furthermore, adversarial examples have cross-models generalization properties \cite{goodfellow2014explaining}, and attackers can generate adversarial examples with attack capability even without knowing the DNN structure \cite{Cnn}.

\par To counter this threat, researchers have proposed many defenses. For example, robust-based defense try to “keep the bad guys out” \cite{das2017keeping}. Many digital image processing methods are adopted, including JPEG compression \cite{das2017keeping}, scaling \cite{lu2017safetynet}, random smoothing \cite{cohen2019certified}, and adding noise \cite{mag2017}. Adversarial training \cite{goodfellow2014explaining,adv} mixes the adversarial examples with normal examples as a training dataset to train models to improve the robustness against specific adversarial attacks. However, many studies have proven that the aforementioned methods can only defend against one or several specific kinds of attacks, but do not work when faced with new or unknown attacks \cite{carlini2017towards}.
\par Therefore, adversarial examples detection technique is proposed to “find the bad guys”, which is called detection-based defense and intended to distinguish between adversarial examples and normal examples. While the detection methods are also subject to well-crafted attacks \cite{carlini2017adversarial}, Aldahdooh et al. \cite{aldahdooh2022adversarial} argue that the detection method might be an added value to the system even if a robust defense classifier is used.

Liang et al. \cite{wangxiaofeng} detect adversarial examples by comparing the classification results of input examples with their denoised versions. Due to the need to guarantee the reliability of denoising results, this contrast detection method is not ideal for detecting attacks with strong perturbations, such as PGD \cite{madry2018towards}. Li et al. \cite{li2017adversarial} demonstrate that the intermediate representation of normal examples in hidden layers of DNN models differs from adversarial examples.  Ye et al. \cite{yefeature} perform adversarial example detection by comparing differences in feature maps. 

\begin{figure*}[!t]
\centering
\includegraphics[width=5 in]{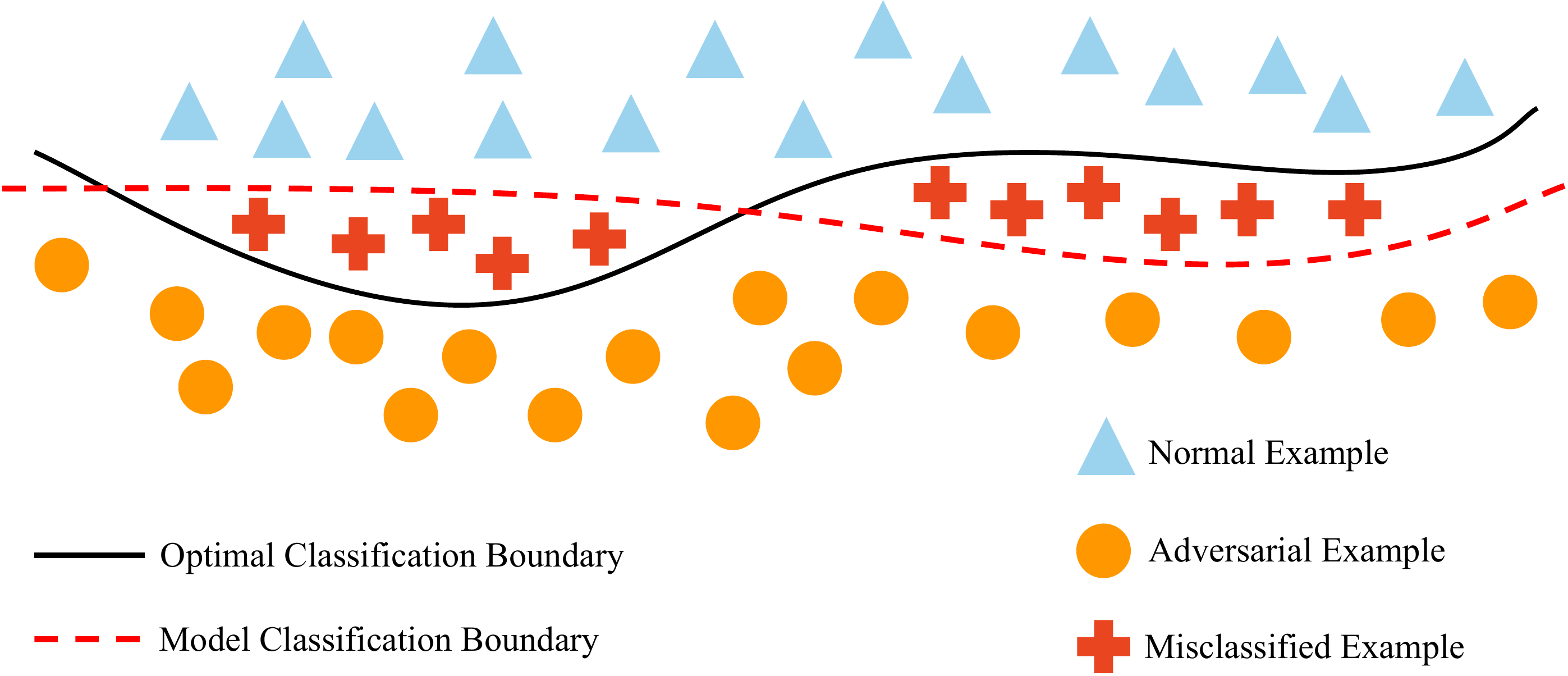}
\caption{Differences between a model classification boundary and the optimal classification boundary.} 
\label{fig1}
\end{figure*}

\par Many adversarial examples detection studies try to find feature differences between normal examples and adversarial examples. Differences between a model classification boundary and the optimal classification boundary of a classifier are shown in Fig. \ref{fig1}. The primary purpose is to bring the detection boundary closer to the optimal boundary. Tian et al. \cite{tian2021detecting} exploit inconsistent sensitivities to detect adversarial examples through a classifier with a transfer decision boundary. To train a binary classifier to distinguish between adversarial examples and normal ones, Agarwal et al. \cite{agarwal2020image} propose to encode the difference between them through image transformation operations and generalized search tree features. 
For the adversarial example detection methods based on searching the feature differences between the adversarial and original examples, the detection performance depends on whether the feature differences found by the method are robust. With advances in attack research, more and more attack methods aim to generate adversarial examples with less perturbation and better visual quality. That makes the feature difference between the adversarial and the original examples too small to be detected. 


During the study, we experimentally observed that the adversarial perturbations existing in the adversarial examples are cluttered in local regions, which is significantly different from the normal examples. Because pixels are often modified in the direction of misleading the target model in the process of generating adversarial examples, the local semantic relevance of the image is not concerned. Therefore, as a kind of high-frequency information in images \cite{Duan_2021_ICCV}, adversarial perturbations in adversarial examples are locally random. In contrast, high-frequency information in normal examples is locally relevant and semantically relevant. Through local histogram equalization, the correlation of local pixels can effectively amplify these abnormal pixels (adversarial perturbation). 


The invisibility against perturbations is one of the evaluation indicators for adversarial example attacks, which is very similar to information hiding \cite{du2020high,yin2020reversible}; This also means that the difference in the local correlation of high-frequency information in adversarial examples and normal examples can be used as an effective feature to distinguish the two. Therefore, we propose an adversarial example detection framework based on a high-frequency information enhancement strategy, which can effectively extract and amplify the feature differences between adversarial examples and normal examples.

In this study, we actively amplify the differences. First, we take advantage of the fact that the adversarial perturbation differs from the original local pixels, and we perform local histogram equalization on each input example. Secondly, high-pass filtering of the processed examples makes the difference in high-frequency information clearer. Finally, it is trained for binary classification. 

The experimental results show that the proposed method can not only detect FGSM \cite{goodfellow2014explaining}, PGD \cite{madry2018towards}, and other gradient-based adversarial examples with a detection accuracy of more than 99\%, but also has a detection accuracy of more than 96\% for optimization-based attacks such as C\&W \cite{carlini2017towards} and DeepFool \cite{moosavi2016deepfool}. Even for UAP \cite{moosavi2017universal}, we can achieve more than 98\% detection accuracy. At the same time, the proposed active enhancement feature difference can improve the accuracy of other detection methods.

\par  The main contributions of this study are summarized as follows:

\begin{itemize}
\item[$\bullet$] We propose an adversarial example detection framework based on high-frequency information enhancement strategy, which can effectively extract and amplify feature differences between adversarial examples and normal examples. 

\item[$\bullet$] The proposed high-frequency information enhancement strategy is general and can be used to improve the performance of all adversarial example detection models.

\item[$\bullet$] Extensive experiments show that the proposed framework not only outperforms the state-of-the-art methods on different attacks and datasets, but also maintains better performance in cross-models detection.
\end{itemize}

\section{Related Work}

\par A brief review of state-of-the-art methods for adversarial attacks and adversarial examples detection is presented in this section. Because there are many methods in these two fields, only some classic and state-of-the-art methods are shown.

\subsection{Adversarial Attack Methods}

\par Szegedy et al. \cite{szegedy2014intriguing} were the first to demonstrate that a neural network can be misled, resulting in misclassification, by adding imperceptible perturbations to images; they proposed a method of generating adversarial examples called box-constrained L-BFGS. The method of generating examples is optimized for the following problem as

\begin{equation}
    \text { Minimize } c|\rho|+\operatorname{loss}_{f}(x+\rho, l) \text { subject to } x+\rho \in[0,1]^{m}
    \label{eq_L-BFGS},
\end{equation}where $f$ represents the trained model, $\rho$ represents adversarial perturbations, and $l$ represents the category that the model is expected to finally predict.

\par Goodfellow et al. \cite{goodfellow2014explaining} proposed a method called FGSM attack. This method has the advantage of high efficiency and does not need to use all the information of the gradient; only the direction of the gradient symbol needs to be determined. Its calculation formula is defined as 
\begin{equation}
x^{a d v}=x+\epsilon \cdot \operatorname{sign}\left(\nabla_{x} J(\theta, x, y)\right),
\label{eq_FGSM}
\end{equation}where $x$ is the original input, $x_{adv}$ is the adversarial example, and $y$ is a fooling category. Perturbations are added to the original input to obtain adversarial examples by the following steps: finding the derivative of the model concerning the input $\nabla_{J}$; using the sign function to get its specific gradient direction $\operatorname{sign}$; and finally multiplying by the step size $\epsilon$.

\par The FGSM method uses gradient information to attack, and the processing speed is relatively fast. However, the FGSM method involves only a single gradient update. Sometimes, a single update is not enough to attack successfully. Therefore, Kurakin et al. \cite{kurakin2017adversarial} introduced an I-FGSM; the specific iteration formula is as follows:

\begin{equation}
   \begin{gathered}
\begin{gathered}
x_{0}^{adv}=x, \\
x_{n+1}^{a d v}=\operatorname{Clip}_{x, \epsilon}\left\{x_{n}^{a d v}-\alpha \operatorname{sign}\left(\nabla_{x} J\left(x_{n}^{a d v}, y_{t}\right)\right)\right\}.
\end{gathered}
\end{gathered}
    \label{eq_I-FGSM}
\end{equation}It operates by adding small perturbations each time, updating the gradient multiple times, and constraining each perturbation to a reasonable area to achieve a better attack effect. 

\par Madry et al. \cite{madry2018towards} proposed a method called PGD attack. PGD attack is an iterative attack. FGSM performs only one iteration and takes a big step. In contrast, PGD performs multiple iterations, taking a small step each time. Each iteration clips perturbations to the specified range. The iteration formula is defined as

\begin{equation}
x_{n+1}^{a d v}=\Pi_{x,\epsilon}\left(x_{n}^{a d v}+\alpha \operatorname{sign}\left(\nabla_{x} J(\theta, x, y)\right)\right),
\end{equation}where $\Pi$($\cdot$) prevents too much perturbation per iteration. The adversarial examples generated by the method are used for adversarial training, which can effectively improve the robustness of models. 

\par DeepFool \cite{moosavi2016deepfool} assumes a linear approximation decision boundary. It changes the classification of examples, by moving them to the linear boundary. The distance from an example to the boundary is where the cost is minimal. The method explores the nearest decision boundary; in each iteration, the image makes minor changes to reach the boundary. C$\&$W \cite{carlini2017towards} is an optimization-based attack method that uses three different metric distances, $L_{0}$, $L_{2}$ and $L_{\infty}$, which generate adversarial examples that are more difficult to discern by the human eye. Moosavi-Dezfooli et al. \cite{moosavi2017universal} proposed a UAP attack method to search for a universal adversarial perturbation that causes the model to misclassify most images.

In addition to the above classic attack methods, some recently proposed complex attack methods balance invisibility and attack, such as adaptive adversarial attacks \cite{wang2019invisible,tramer2020adaptive}, semantic segmentation-based attack \cite{li2019adversarial}, and soon.
 
\subsection{Robust-based Defense}

\par Robust-based defenses aim to correctly classify adversarial examples. Currently, there are many ways to implement an effective defense. Adversarial training \cite{szegedy2014intriguing,adv} mixes adversarial examples into normal examples as a training set to train a more robust network model. Preprocessing is conducted to perform auxiliary operations on input examples, including principal component analysis \cite{li2017adversarial}, JPEG compression \cite{das2017keeping}, noise addition \cite{deng2021detecting}, cropping \cite{song2021learning}, rotation \cite{song2021learning}, etc. Wei et al. \cite{wei} proposed an ensemble defense method that combines multiple image denoising techniques. Zheng et al. \cite{grip} proposed GRIP-GAN to generate a generalized robust inverse perturbation (GRIP), which not only counteracts potential adversarial perturbations in input examples but also enhances the target class-related features of normal examples.

\subsection{Detection-based Defense}

\par Detection-based defenses aim to distinguish between normal and adversarial examples. Hendrycks et al. \cite{hendrycks2017baseline} found that correctly classified examples have a larger maximum softmax probability than misclassified and out-of-distribution examples. Liang et al. \cite{liang2017principled} proposed a detector to use temperature scaling and input processing to improve the model's out-of-distribution detection ability. Liang et al. \cite{wangxiaofeng} treated adversarial attacks as noise in an input image. They used quantization and image filtering techniques in the spatial domain to mitigate the effects of adversarial noise and detect adversarial examples by comparing the classification results before and after processing. Goswami et al. \cite{goswami2019detecting} proved that the intermediate representation of the hidden layer of a DNN model of normal examples was different from adversarial examples and performed adversarial detection by evaluating the different behavior of the hidden layers. 
Feinman et al. \cite{feinman2017detecting} proposed to detect adversarial examples by referring to Bayesian uncertainty estimates and density estimates represented in the deep feature subspace learned by a model. Liu et al. \cite{liu2019detection} modeled the differences between adjacent pixels in normal examples to identify biases caused by adversarial attacks. Adversarial examples are detected using a detector to detect transition probability matrices by modeling dependencies between adjacent pixels of the filtered image. Then the transition probability matrix was used as a vector feature for feature-based detectors. Agarwal et al. \cite{agarwal2020image} proposed an adversarial examples detection method based on image transformation, which used image-discrete cosine transformation and discrete wavelet transformation to encode edge information by calculating multi-scale and multi-directional gradients. Tian et al. \cite{tian2021detecting} found that normal examples were insensitive to fluctuations in the curved region of the decision boundary.
In contrast, adversarial examples were overly sensitive to such fluctuations. They designed a  detector with a transformed decision boundary to take advantage of the inconsistencies in the sensitivity of different examples to the decision boundary to detect adversarial examples. Ye et al. \cite{yefeature} encoded and reconstructed an input example's feature maps and detected them by comparing the difference between them before and after reconstruction.

\section{Proposed Method}

\par While humans cannot be confused by adversarial examples, it is difficult for machines to classify adversarial examples correctly. We found that the difference between the two is tiny at the pixel level. At the model feature level, the difference is slightly amplified. When these differences are significant enough, the detector can accurately distinguish which are adversarial examples.

\par Adversarial perturbations and image textures in adversarial examples are high-frequency information \cite{Duan_2021_ICCV}. However, adversarial perturbations are independent of images. In contrast, image textures are locally correlated, and these irrelevant adversarial perturbations turn into anomalous information. Enhancing image contrast amplifies the difference between normal and adversarial examples to help a detector perform better. 

\par As shown in Fig. \ref{fig2}, the proposed framework is divided into three parts. The first part is the image contrast enhancement module, which improves the contrast of high-frequency information. The second part is the high-frequency information filter module, which extracts the enhanced adversarial features. The third part is the neural network classification module. The image is input into the neural network for binary classification training. First, the image contrast enhancement module performs RGB local histogram equalization on input examples according to the pre-selected sliding window size. (The sliding window size is given in Sec. \ref{sec:Ablation study on Filter}). The high-frequency filtering module filters through the second-order Butterworth high-pass filter. Finally, the filtered high-frequency information feature maps are trained for binary classification.
\begin{figure*}[!t]
    \centering
    \includegraphics[width = 6.5 in]{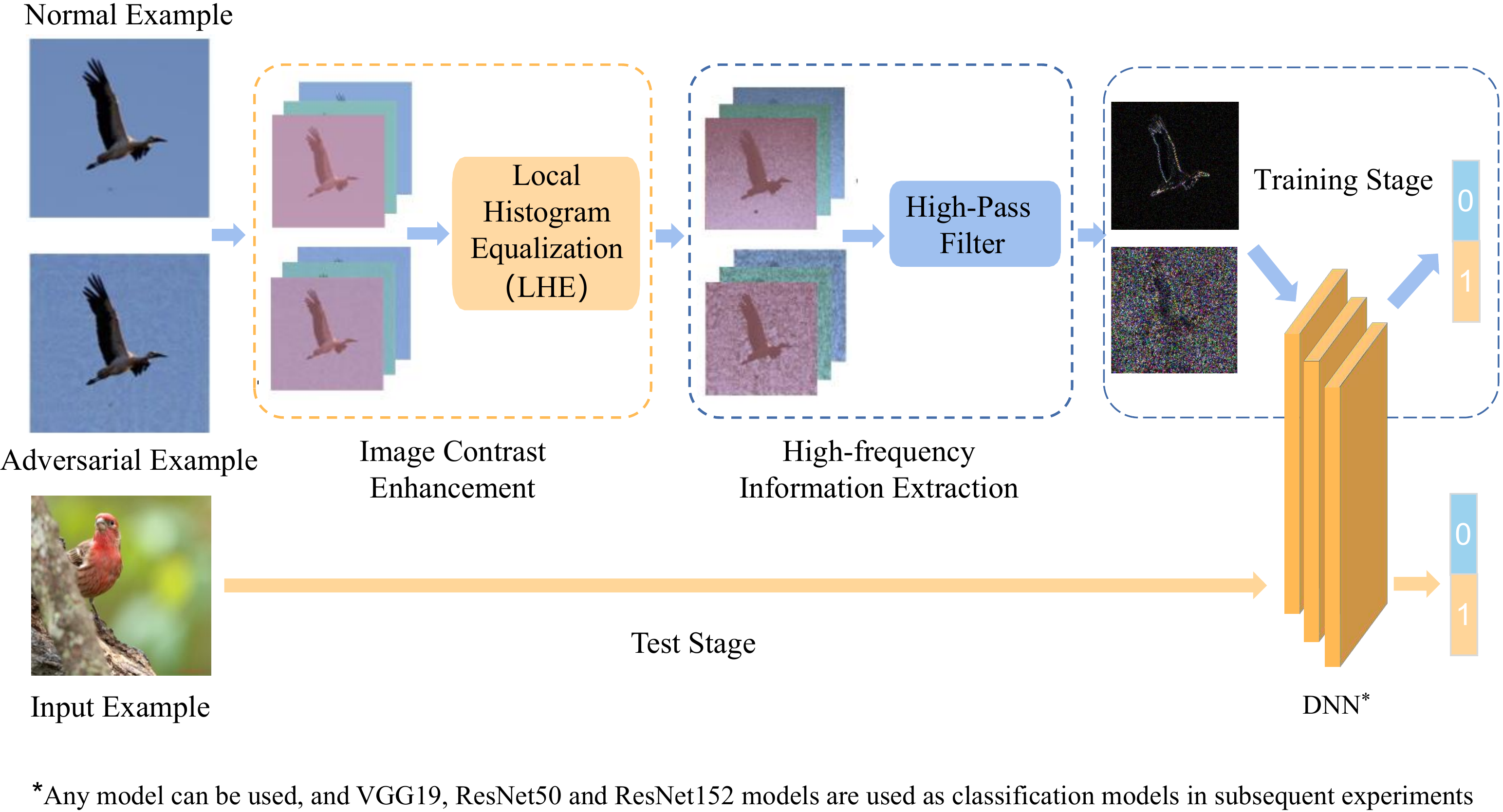}
    \centering
    \caption{\centering Framework diagram of the proposed detection method.}
    \label{fig2}
\end{figure*}

\subsection{Problem Formulation}

Based on adversarial detection methods that identify adversarial features, these methods judge whether an example is adversarial by training a model to identify whether an example contains adversarial features. Most detection methods focus on finding more powerful adversarial features.

\subsubsection{Why local histogram equalization?}

At present, research on adversarial example attacks is devoted to generating low-perturbation, highly aggressive adversarial examples. The adversarial features in future adversarial examples are more stealthy and difficult to detect. This means that adversarial detection methods that rely only on adversarial features in adversarial examples become less reliable. Therefore, it would be better to actively amplify the abnormal features (adversarial features) in the adversarial examples. Taking advantage of the randomness of the adversarial perturbation in the local area of the image, we use the method of local histogram equalization to make the image more recognizable.

\subsubsection{Local Histogram Equalization and High-Pass Filter}

It is well known that adversarial perturbations and complex textures in images are high-frequency information. Although the image after local histogram equalization already has relatively recognizable features. However, for images with complex image textures, there will still be more non-adversarial high-frequency information in the image after local histogram equalization. We use the high-pass filter to retain more adversarial high-frequency information (adversarial perturbation) at an appropriate cut-off frequency, enabling the classifier to obtain better classification performance. More details are given in Sec. \ref{sec:High-Frequency Information based on LHE} and Sec. \ref{sec:High-Frequency Information Extraction}.

\subsection{High-Frequency Information based on LHE  \label{sec:High-Frequency Information based on LHE}}

\par When humans identify image categories, they pay more attention to the contours of the image than to the texture details in the image. Deep neural networks focus too much on high-frequency texture information on images. However, adversarial perturbations are also high-frequency information \cite{Duan_2021_ICCV}, which means that these adversarial perturbations easily mislead deep neural networks. How to improve DNN's discrimination between high-frequency texture information and adversarial perturbations? Some additional operations can be used to enhance the ability of deep neural networks to identify unnatural high-frequency information.

\par In digital image processing, the histogram is a vital image feature. The histogram reflects the statistical properties of the image pixel distribution. There are many histogram equalization methods for the contrast enhancement of digital images. Global histogram equalization (GHE) is a common histogram equalization method. The basic idea of the GHE method is to remap the gray level of images according to the gray cumulative density function. Although GHE can effectively extend the dynamic range of digital images, the processed histograms are not smooth. The histogram of output images contains many empty intervals because GHE performs global enhancement without considering the image's local correlation. An extension called local histogram equalization (LHE) was introduced to solve the problem. LHE performs a transformation function based on the adjacent pixels around each pixel, using a sliding window to equalize the entire image locally. The LHE image processing process is shown in Fig. \ref{fig3}.

The image processed by LHE can be passed to the high-frequency information extraction module to obtain more apparently abnormal pixels (adversarial perturbation).

\begin{figure}[!t]
    \centering
    \includegraphics[width = 2.5 in]{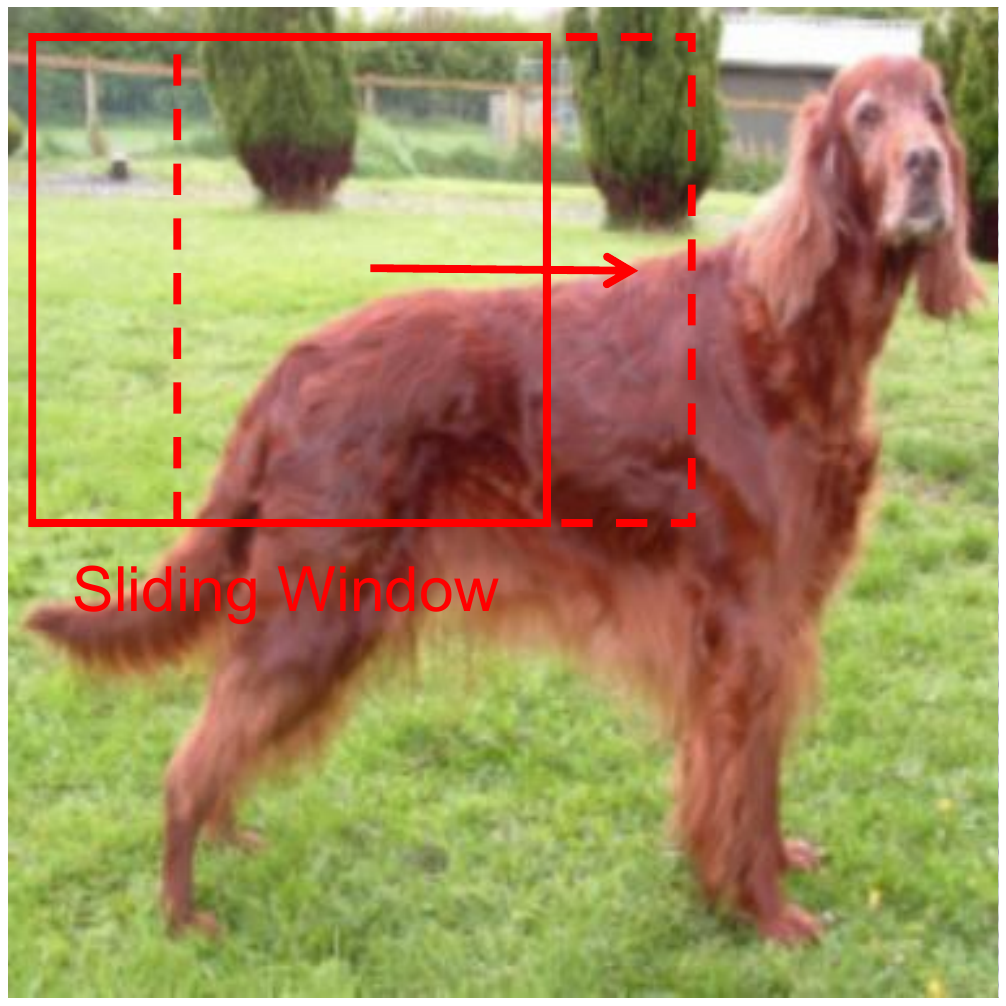}
    \caption{ LHE is achieved by continuously sliding the processing window.}
    \label{fig3}
\end{figure}

\subsection{High-Frequency Information Extraction \label{sec:High-Frequency Information Extraction}}
We introduce several common filters in this section and analyze their characteristics. 

\subsubsection{High-Pass Filter}

\par The high-pass filter blocks the signal below a threshold by setting a threshold. However, when processing digital images, the high-pass filter abruptly cuts off the signal at the set cut-off frequency, resulting in less smooth edges of the filtered high-frequency information. The transfer function of the high-pass filter is defined as
\begin{equation}
    H(u, v)= \begin{cases}0 & D(u, v) \leq D_{0} \\ 1 & D(u, v)>D_{0}\end{cases} ,
\end{equation}
$D_{0}$ is the cut-off radius of the filter, $D(u, v)$ is the distance from the pixel to the center of the input image (Euclidean distance), and the calculation formula is as follows:

\begin{equation}
    \mathrm{D}(u, v)=\sqrt{(u-M / 2)^{2}+(v-N / 2)^{2}},
\end{equation}where $M$ and $N$ represent the size of an input image.

\subsubsection{Gaussian High-Pass Filter}

\par The Gaussian high-pass filter is a linear smoothing filter. The main idea is the weighted average method, which gives the center pixel the maximum weight and considers the surrounding pixels. The weight mainly depends on how close the surrounding pixels are to the center pixel. The transfer function of the Gaussian high-pass filter is defined as
\begin{equation}
    H(u, v)=1-e^{-D^{2}(u, v) / 2 D_{0}{ }^{2}} .
\end{equation}

\subsubsection{Butterworth High-Pass Filter}

\par The Butterworth high-pass filter is also widely used. The Butterworth filter has the flattest amplitude response in the passband and has no ripple. $D_{0}$ represents the radius of the passband, and $n$ represents the order of the Butterworth filter. As the order increases, the ringing phenomenon becomes more pronounced. The expression is 
\begin{equation}
    H(u, v)=\frac{1}{1+\left[D_{0} / D(u, v)\right]^{2 n}} .
\end{equation}
The selection of the high-pass filter and its parameters is explained in detail in the ablation studies section (Sec. \ref{sec:Ablation study on Filter}).

\subsection{Training Detector}

\par After enhancing the image contrast, the filtered high-frequency information is sent to the classification model for two-class training. The trained detector can be used to detect other adversarial examples.

\par Enhancing the differences in image features requires training a binary classification detector, implemented using a preprocessed example set and default settings (ResNet50 \cite{he2016deep}). The trained ResNet50 \cite{he2016deep} model, for images of different sizes, MNIST \cite{lecun1998gradient}, CIFAR-10 \cite{krizhevsky2009learning}, and ImageNet \cite{krizhevsky2012imagenet} all have satisfactory classification accuracy.

\section{Experimental Results}

\par To evaluate the proposed method and compare with classical and state-of-the-art attack detection methods (ITGS \cite{agarwal2020image}, Adaptive Noise \cite{wangxiaofeng}, SID \cite{tian2021detecting}, Bayesian Uncertainty \cite{feinman2017detecting}, Base-OOD \cite{hendrycks2017baseline}, ODIN \cite {liang2017principled}, ESRM \cite{liu2019detection}, and FADetector \cite{yefeature}), different adversarial attacks(FGSM \cite{goodfellow2014explaining}, I-FGSM \cite{kurakin2017adversarial}, PGD \cite{madry2018towards}, DeepFool \cite{moosavi2016deepfool}, C\&W \cite{carlini2017towards} and UAP \cite{moosavi2017universal}) are applied to generate adversarial examples for three image classification models (VGG19 \cite{Simonyan2015vgg}, ResNet50 \cite{he2016deep} and ResNet152 \cite{he2016deep}) on three commonly used datasets (MNIST \cite{lecun1998gradient}, CIFAR-10 \cite{krizhevsky2009learning} and ImageNet \cite{krizhevsky2012imagenet}).

\par In this section, we first compare the performance with classical and state-of-the-art detection methods on different datasets. Then, the cross-models detection performance experiments of the proposed method are carried out, and the problems in the experiments are analyzed and solved. Finally, ablation studies are performed between different modules, and the influence of different modules on the overall detection process is analyzed.

\subsection{Implementation Details}

\par The PyTorch-based attack library function is used in the experiment, which can directly generate various adversarial examples. The parameters of various attack methods are shown in Table \ref{tab:1}. 2000 examples were randomly selected and trained using ResNet50 \cite{he2016deep} on the MNIST \cite{lecun1998gradient} and CIFAR-10 \cite{krizhevsky2009learning} datasets. 10000 raw examples are randomly selected on the ImageNet dataset and trained on ResNet50 \cite{he2016deep}, ResNet152 \cite{he2016deep}, and VGG19 \cite{Simonyan2015vgg}.

\begin{table}[!h]
	\centering
	\caption{Parameters for generating adversarial examples.}
    \label{tab:1}
	{
	\begin{tabular}{m{1.8cm}<{\raggedright}m{5cm}<{\raggedright}}
		\toprule
		\textbf{Attacks}&\textbf{Parameters}\\
		\midrule
		FGSM \cite{goodfellow2014explaining}& eps = 0.3, norm = $L_{\infty}$, b$\_$size = 128\\
		PGD \cite{madry2018towards}& eps = 0.3, norm = $L_{\infty}$, max$\_$iter = 100\\
		DeepFool \cite{moosavi2016deepfool}& max$\_$iter = 100, b$\_$size = 32\\
		C\&W \cite{carlini2017towards}& b$\_$size = 32, max$\_$iter = 100, k = 20\\
		UAP \cite{moosavi2017universal}& max$\_$iter = 1000, b$\_$size = 64 \\
		\bottomrule
	\end{tabular}
    }
\end{table}

\subsection{Adversarial Detection Accuracy Comparison on MNIST and CIFAR-10 Datasets}

\begin{figure*}[!t]
    \centering
    \includegraphics[width = 5 in]{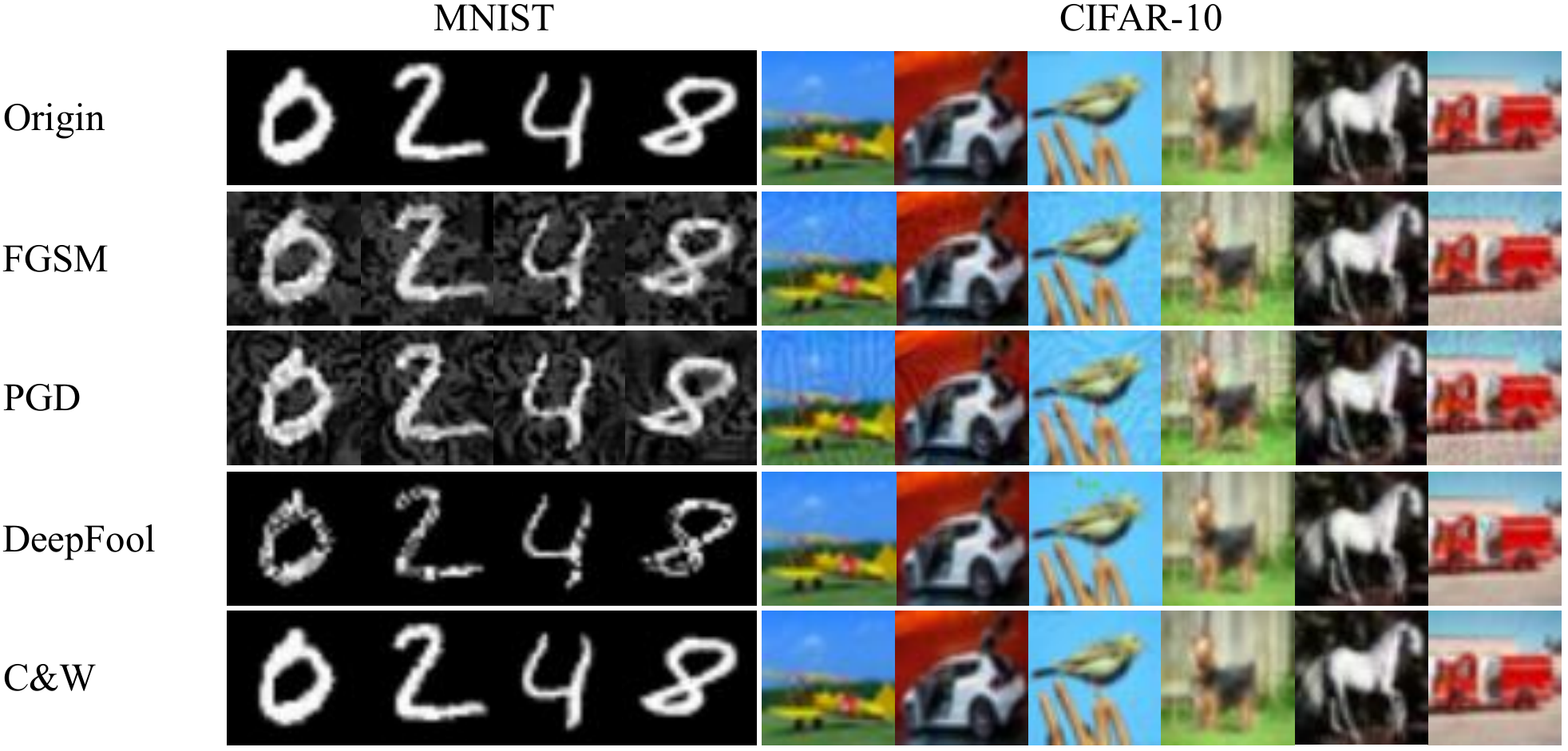}
    \caption{Four types of adversarial examples in MNIST and CIFAR‐10 with high attack success rate. }
    \label{fig8}
\end{figure*}

\par The generated adversarial examples are shown in Fig. \ref{fig8}, and the recognition accuracy of the ResNet50 \cite{he2016deep} model for these adversarial examples is shown in Table \ref{tab:2}, where the accuracy of four different mainstream attacks (FGSM \cite{goodfellow2014explaining} ,PGD \cite{madry2018towards}, DeepFool \cite{moosavi2016deepfool} and C\&W \cite{carlini2017towards}) are evaluated. Table \ref{tab:3} summarizes the detection performance of the proposed method and existing detection methods. It can be seen from the comparison of the results that the detection accuracy of the existing method for the strongest attack PGD \cite{madry2018towards} is 99.4\% by SID \cite{tian2021detecting}, while that of the proposed methed is 99.9\%. Attacks based on FGSM \cite{goodfellow2014explaining} and PGD \cite{madry2018towards} are almost entirely detected. Other attacks also can be detected with high probability, and the proposed method works best. Another sophisticated attack, DeepFool \cite{moosavi2016deepfool}, has a detection accuracy of 97.2\%, which is 0.9\% higher than existing boundary inconsistency-based detection methods. Although the performance of the proposed method in C\&W \cite{carlini2017towards} attack detection is about 0.1\% lower than that of SID \cite{tian2021detecting}, that is because SID \cite{tian2021detecting} exploits the fluctuation of the classification boundary for detection, which has an advantage over adversarial examples with less perturbation intensity. Compared with other methods, the proposed method has great advantages for adversarial examples at almost all strengths. 

\begin{table}[!t]
	\centering
	\caption{Accuracy (\%) under different attacks on MNIST and CIFAR-10 datasets.}
    \label{tab:2}
	
	{
	\begin{tabular}{m{1.8cm}<{\raggedright}m{2cm}<{\centering}m{2cm}<{\centering}}
		\toprule
		\textbf{Attack}&\textbf{MNIST} &\textbf{CIFAR-10}\\
		\midrule
		No attack& 99.3&87.6\\
		FGSM \cite{goodfellow2014explaining}& 21.6&10.1\\

		PGD \cite{madry2018towards} & 0.6&1.3\\
	
		 DeepFool \cite{moosavi2016deepfool}& 13.6&36.5\\

		 C\&W \cite{carlini2017towards}& 17.2&40.5\\
		\bottomrule
	\end{tabular}
    }
\end{table}

\begin{table}[!h]
    \centering
    \caption{Adversarial examples detection accuracy (\%) of the proposed and existing methods on the MNIST dataset.}
    \label{tab:3}
    \begin{tabular}{m{2.5cm}<{\raggedright}m{0.9cm}<{\centering}m{0.7cm}<{\centering}m{1.4cm}<{\centering}m{0.8cm}<{\centering}}

    \toprule
    \textbf{Attack} & \textbf{FGSM} & \textbf{PGD}  & \textbf{DeepFool} & \textbf{C\&W} \\
    \midrule
    AdaptiveNoise \cite{wangxiaofeng} & 92.8 & 74.6 & 72.1 & 79.2 \\
    BU \cite{feinman2017detecting} & 82.2 & 77.8 & 72.1 & 79.2 \\
    SID \cite{tian2021detecting} & 91.9 & 99.4 & 96.3 & \textbf{96.6} \\
    FADetector \cite{yefeature} & 75.3 & 73.2 & 81.2 & 74.7 \\
    ESRM \cite{liu2019detection} & 74.3 & 71.5 & 81.1 & 67.5 \\
    ITGS \cite{agarwal2020image} & 98.3 & 97.8 & 68.4 & 96.4 \\
    Proposed & \textbf{99.9} & \textbf{99.9} & \textbf{97.2} &96.5\\
    \bottomrule
    \end{tabular}
\end{table}

\par Since the C\&W \cite{carlini2017towards} and DeepFool \cite{moosavi2016deepfool} attack methods require repeated iterative attacks on input examples to generate imperceptible perturbations, they are more difficult to detect, with the result that the attack success rate is not as high as that of the FGSM \cite{goodfellow2014explaining} and PGD \cite{madry2018towards} attack methods.

\par Table \ref{tab:4} shows the comparison results of the proposed method with other detection methods on the CIFAR-10 dataset \cite{krizhevsky2009learning}. The proposed method detects PGD \cite{madry2018towards} attacks with 97.7\% accuracy. The PGD \cite{madry2018towards} attack method is the strongest first-order attack method. The high detection accuracy of PGD \cite{madry2018towards} proves that the proposed method is effective for complex attack methods.

\begin{table}[!h]
    \centering
    \caption{Adversarial examples detection accuracy (\%) of the proposed and existing methods on the CIFAR-10 dataset.}
    \label{tab:4}
    \begin{tabular}{m{2.5cm}<{\raggedright}m{0.9cm}<{\centering}m{0.7cm}<{\centering}m{1.4cm}<{\centering}m{0.8cm}<{\centering}}

    \toprule
    \textbf{Attack} & \textbf{FGSM} & \textbf{PGD}  & \textbf{DeepFool} & \textbf{C\&W} \\
    \midrule
    AdaptiveNoise \cite{wangxiaofeng} & 83.2 & 59.2 & 57.2 & 57.8 \\
    BU \cite{feinman2017detecting} & 84.0 & 56.5 & 58.6 & 57.3 \\
    SID \cite{tian2021detecting} & 87.5 & 93.6 & 85.3 & 91.5 \\
    FADetector \cite{yefeature} & 73.4 & 74.4 & 73.5 & 80.8 \\
    ESRM \cite{liu2019detection} & 61.4 & 66.4 & 69.3 & 67.5 \\
    ITGS \cite{agarwal2020image} & 93.4 & 93.6 & 73.2 & 74.4 \\
    Proposed & \textbf{97.5} & \textbf{97.7} & \textbf{96.8} & \textbf{94.3}\\
    \bottomrule
    \end{tabular}
\end{table}

\par Likewise, in the DeepFool \cite{moosavi2016deepfool} attack, the detection performance is at least 96.8\%, which is 23.3\% better than FADetector \cite{yefeature} and 11.5\% better than SID \cite{tian2021detecting}. Furthermore, the proposed method outperforms the ESRM \cite{liu2019detection} method by 25.8\% on the C\&W \cite{carlini2017towards} attack detection results.

\subsection{Adversarial Detection Accuracy Comparison on ImageNet Dataset}

\par This experiment uses 10,000 original images, and 8485 adversarial examples were generated using FGSM \cite{goodfellow2014explaining}, I-FGSM \cite{kurakin2017adversarial}, PGD \cite{madry2018towards}, and UAP \cite{moosavi2017universal} attack methods. The ResNet50 \cite{he2016deep} and VGG16 \cite{Simonyan2015vgg} models generate the universal adversarial examples, reducing the model accuracy to 6.3\%, with a high attack success rate. Fig. \ref{fig9} shows the generated universal perturbations and universal adversarial examples. The LHE sliding window size in the detection method is set to 27, and the filter cut-off frequency is 56. We compare the performance with several current state-of-the-art methods (better value selected by \ref{sec:Ablation study on Filter} \nameref{sec:Ablation study on Filter}). As seen in Fig. \ref{fig10}, the proposed method can achieve a detection success rate of 98.2\%; that is because the proposed method first enhances the contrast of local pixels. The local pixels in normal examples are random, and each example is different. At the same time, the universal adversarial perturbation in universal adversarial examples is relatively fixed, and most examples have similar high-frequency information. Therefore, the proposed method has a significant advantage in detecting universal adversarial perturbations.

\begin{table*}[!b]
	\centering
	\caption{Cross-models adversarial example detection accuracy (\%) on the ImageNet dataset.}
    \label{tab:5}

	\begin{tabular}{ccccccccccc}
		\toprule
		& Attacked Model& &VGG19& & &ResNet50& &  &ResNet152& \\
		\cmidrule(r){3-5}\cmidrule(r){6-8}\cmidrule(r){9-11}
        Detection Model& Attack Method& FGSM& I-FGSM& PGD& FGSM& I-FGSM& PGD& FGSM& I-FGSM& PGD \\
        \midrule
        & FGSM&\textbf{99.1}& 99.0& 99.3& 99.0& 49.3& 99.4& 95.0& 55.6& 97.1\\
        VGG19& I-FGSM& 94.3& {\textbf{99.1}}& 99.2& {98.7}& {74.7}& 99.3& {90.5}& {53.4}& 98.9 \\
        & PGD & {89.1}& {93.0}& \textbf{99.5} & {99.3} & {58.8} & 99.4 & {80.3} & {50.4} & 88.4  \\
        \cmidrule(r){2-11}
        & FGSM &{88.1} &{83.9} & 82.7& {\textbf{99.5}}& {62.4}& 96.2& {92.5}& {64.7}& 87.7  \\
        ResNet50& I-FGSM &{91.0}&97.2& 96.5& 96.2&\textbf{98.3}& 97.2&92.3& 90.9& 97.7  \\
        & PGD& {88.3}& {90.1}& 99.2& {99.4} & {53.8}& \textbf{99.5}& {83.4}& {62.2} & 92.1 \\
         \cmidrule(r){2-11}
         & FGSM& 99.4 & 99.1 & 99.8 & 99.9 & 81.1& 99.8 & \textbf{99.9}& 78.6 & 97.5  \\
          ResNet152& I-FGSM & 91.7& 96.6 & 94.3 & 99.6 & 99.3 & 94.9 & 93.8 &\textbf{98.7}& 98.4 \\
          & PGD & 94.9&99.8 & 99.9 &99.9 &56.7 & 99.9&94.2& 61.0  & \textbf{99.9} \\
		\bottomrule
	\end{tabular}
	\centering
\end{table*}

\begin{figure}[!h]
    \centering
    \includegraphics[width = 3.5in]{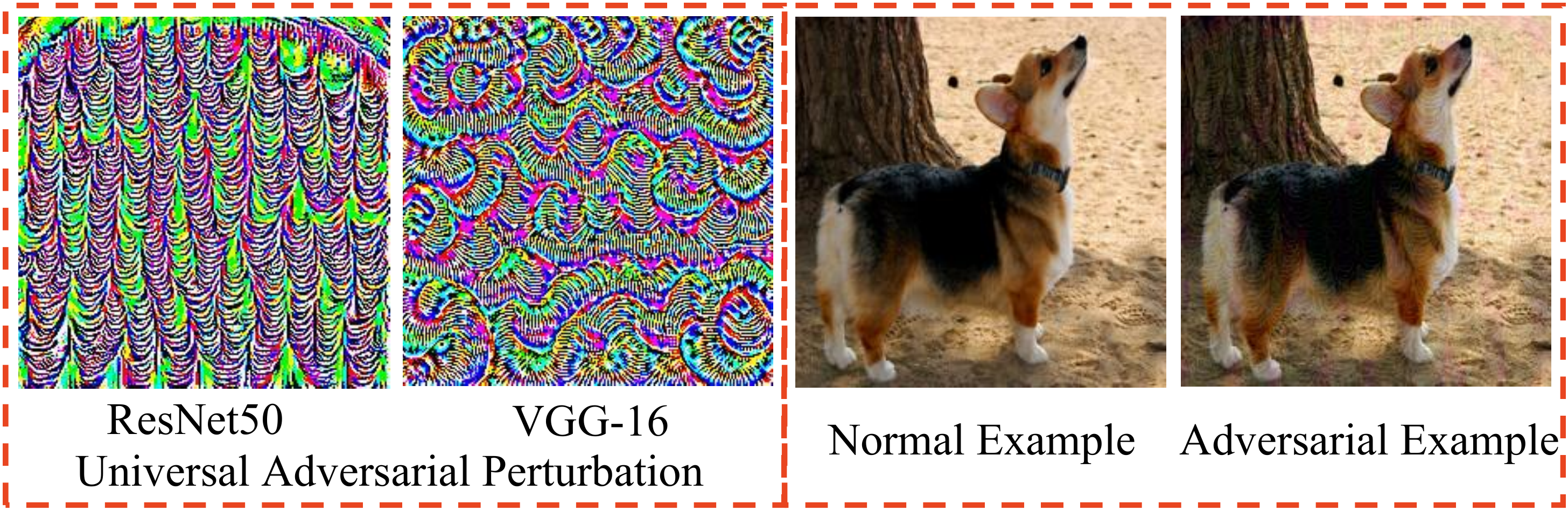}
    \caption{Adversarial examples with addition of universal adversarial perturbation with ResNet50 and VGG-16 architectures. }
    \label{fig9}
\end{figure}

\begin{figure}[!h]
    \centering
    \includegraphics[width = 3.3 in]{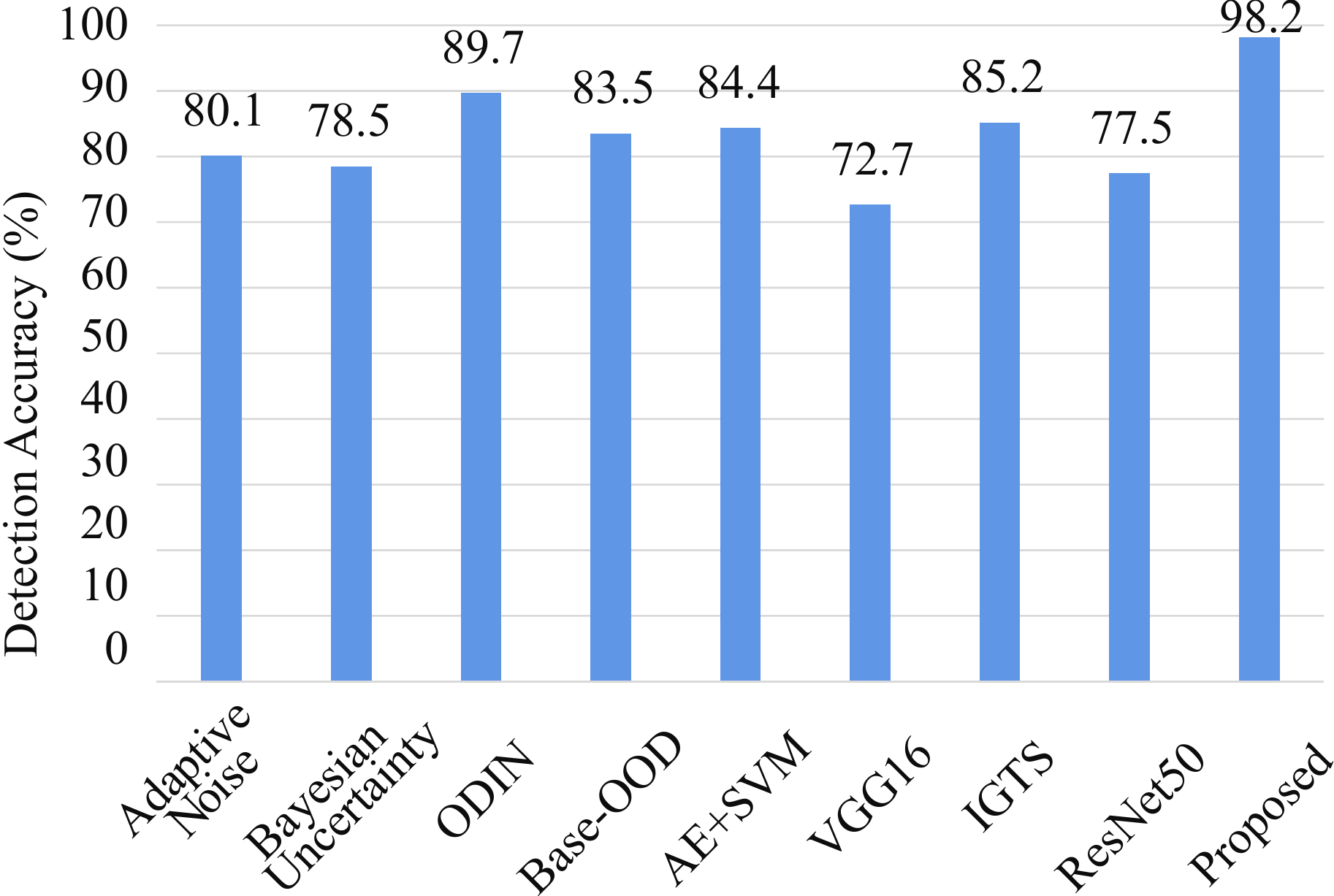}
    \caption{Detection performance of the proposed method and existing adversarial examples detection methods on the ImageNet database with universal adversarial perturbation. }
    \label{fig10}
\end{figure}

\subsection{Adversarial Detection Accuracy Analysis on Cross-Models}

\par At the same time, the experiment uses three models of ResNet50 \cite{he2016deep}, ResNet152 \cite{he2016deep}, VGG19 \cite{Simonyan2015vgg}, and three attack methods  (FGSM \cite{goodfellow2014explaining}, I-FGSM \cite{kurakin2017adversarial} and PGD \cite{madry2018towards}) to test the performance of adversarial example detection. The results shown in bold face are the detection performance of the detector that was trained using the adversarial examples generated by the same model. The other parts can be regarded as the detection performance across models. The test results are shown in Table \ref{tab:5}. It can be seen from the results that, among the three models, all non-cross-models adversarial example detection results can achieve almost a 99\% detection success rate. On the ResNet152 \cite{he2016deep} model, the PGD \cite{madry2018towards} attack detection success rate reaches 99.9\%. In cross-models adversarial example detection, the success rate of I-FGSM \cite{kurakin2017adversarial} attack detection has decreased. This paper conducts relevant experiments and analyzes the reasons for this situation.

\begin{table}[!h]
	\centering
	\caption{The magnitude of different perturbations under the $L_{2}$-norm.}
    \label{tab:6}
    {
	\begin{tabular}{m{2cm}<{\centering}m{1.5cm}<{\centering}m{1.5cm}<{\centering}m{1.5cm}<{\centering}}
	\toprule
	        & FGSM & I-FGSM & PGD \\
	\midrule
	    VGG19 ($L_{2}$)& 11.3966& 4.4233 & 8.5022  \\ 

        ResNet50 ($L_{2}$)& 11.4415& 4.5074& 8.4853 \\ 
  
        ResNet152 ($L_{2}$) & 11.4416& 4.5036& 8.4835  \\ 
	\bottomrule
	\end{tabular}
	}
	\centering
\end{table}

\par In Table \ref{tab:6}, by evaluating the perturbation intensity generated by each attack method under the $L_{2}$-norm, it can be found that, under the same $\epsilon$, FGSM produces the highest perturbation intensity, followed by PGD \cite{madry2018towards}. The perturbation intensity of I-FGSM \cite{kurakin2017adversarial} is very small, so the classification boundary trained using the adversarial examples generated by I-FGSM \cite{kurakin2017adversarial} is more robust to larger perturbations.

\begin{figure}[!h]
    \centering
    \includegraphics[width = 3.5 in]{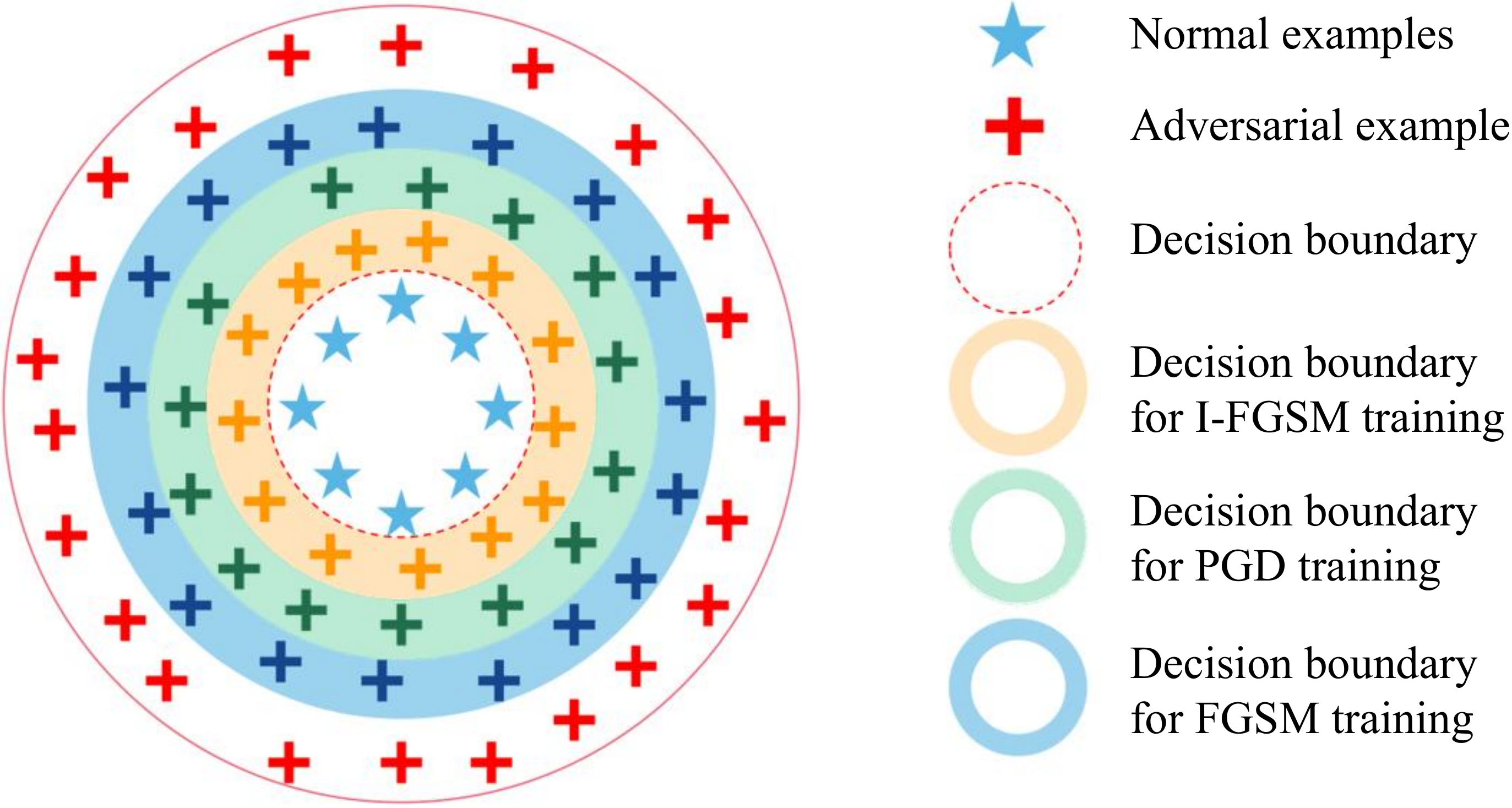}
    \caption{The detection boundaries of the models trained by examples generated by different attack methods are different — the greater the perturbation, the larger the detection boundaries of the models trained by the examples. }
    \label{fig11}
\end{figure}

Specifically, as shown in Fig. \ref{fig11}, the innermost boundary can be regarded as the classification boundary between adversarial examples and normal examples, and each outer circle corresponds to the boundary trained by different adversarial examples. The closer to the inner circle, the better the robustness. This may explain why the detector performs poorly on I-FGSM \cite{kurakin2017adversarial} in terms of cross-models attack detection.

The features of adversarial examples with large perturbations are farther away from the decision boundary. In contrast, the decision boundaries of detectors trained with small perturbations are closer to the classification boundaries of normal examples, resulting in better detection results, and vice versa.

The experimental results in Table \ref{tab:5} show that the classification models with different structures can maintain high detection performance, which indicates that the proposed method does not depend on the specific classification model structure. That means that even if the defender uses the “worst” network structure in actual deployments, it will not cause a significant drop in detection performance.

\begin{figure*}[!ht]
    \centering
    \subfigure[ Pixel-level without LHE]{
        \centering
        \begin{minipage}[b]{0.48\textwidth}
            \includegraphics[width=1\textwidth]{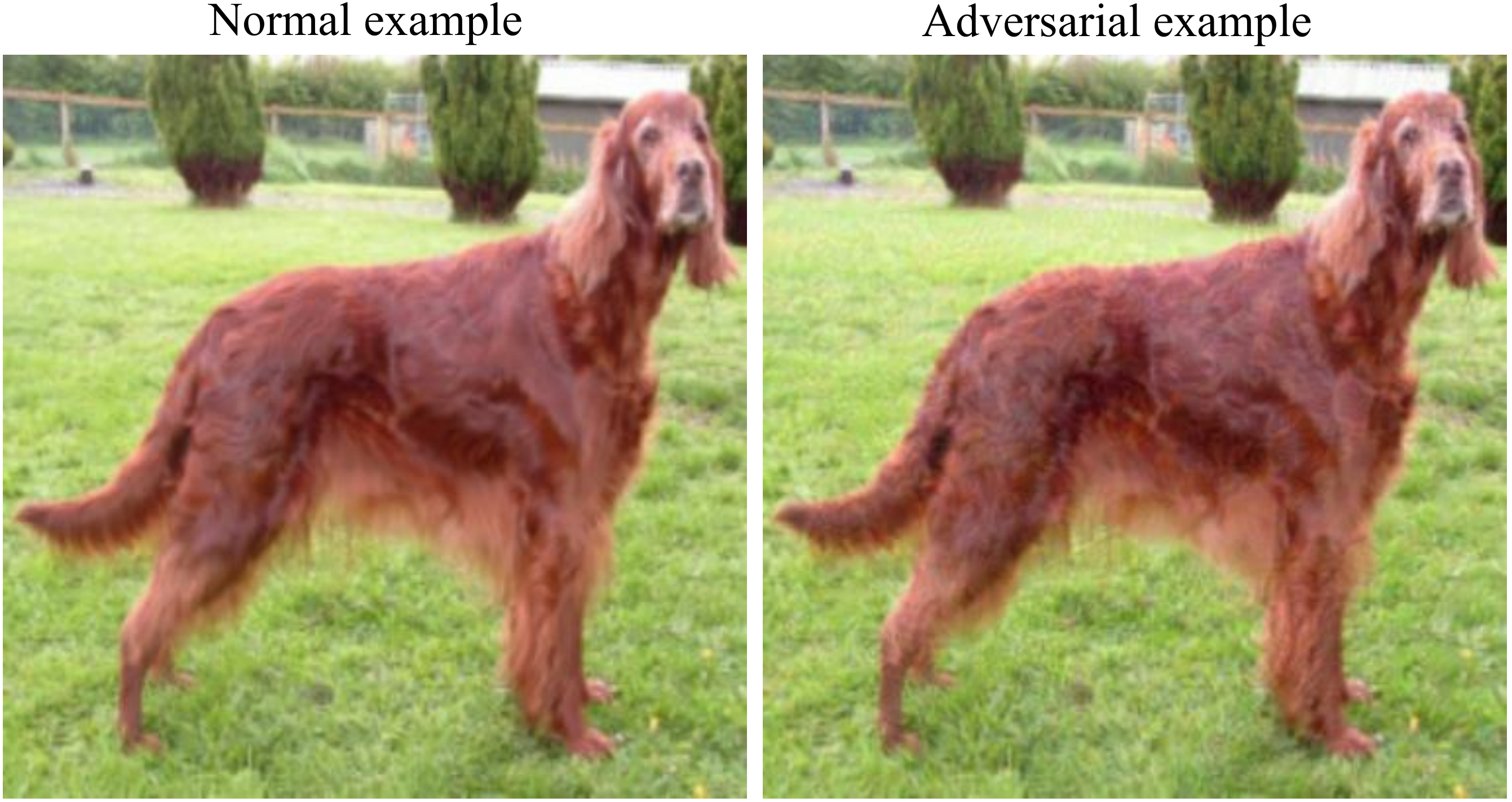} 
        \label{a}
        \end{minipage}
    }
    \subfigure[Feature-level without LHE]{
        \centering
        \begin{minipage}[b]{0.48\textwidth}
            \includegraphics[width=1\textwidth]{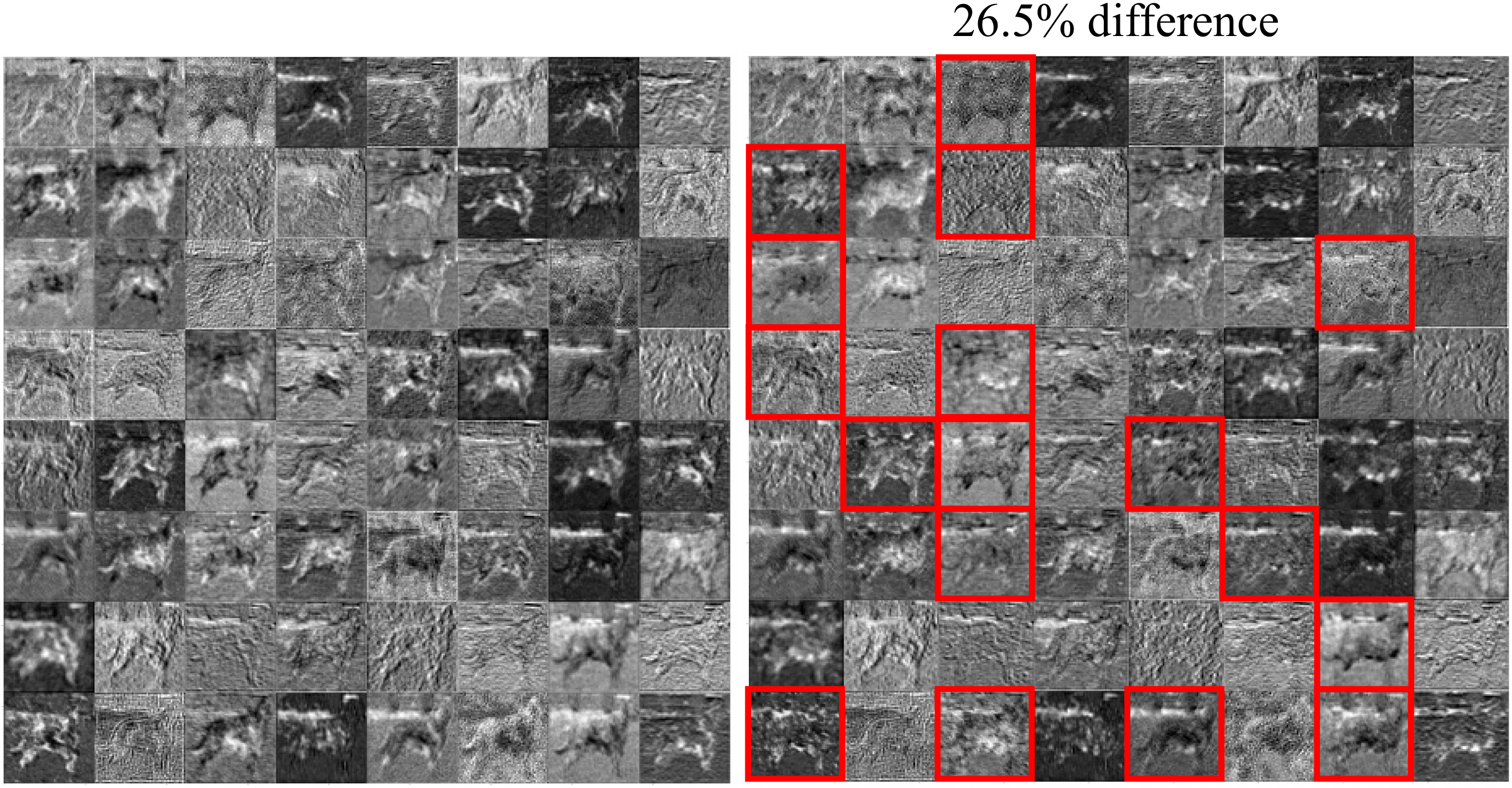} 
            \label{b}
        \end{minipage}
    }
    \quad

    \centering
    \subfigure[Pixel-level with LHE]{
        \centering
        \begin{minipage}[b]{0.48\textwidth}
            \includegraphics[width=1\textwidth]{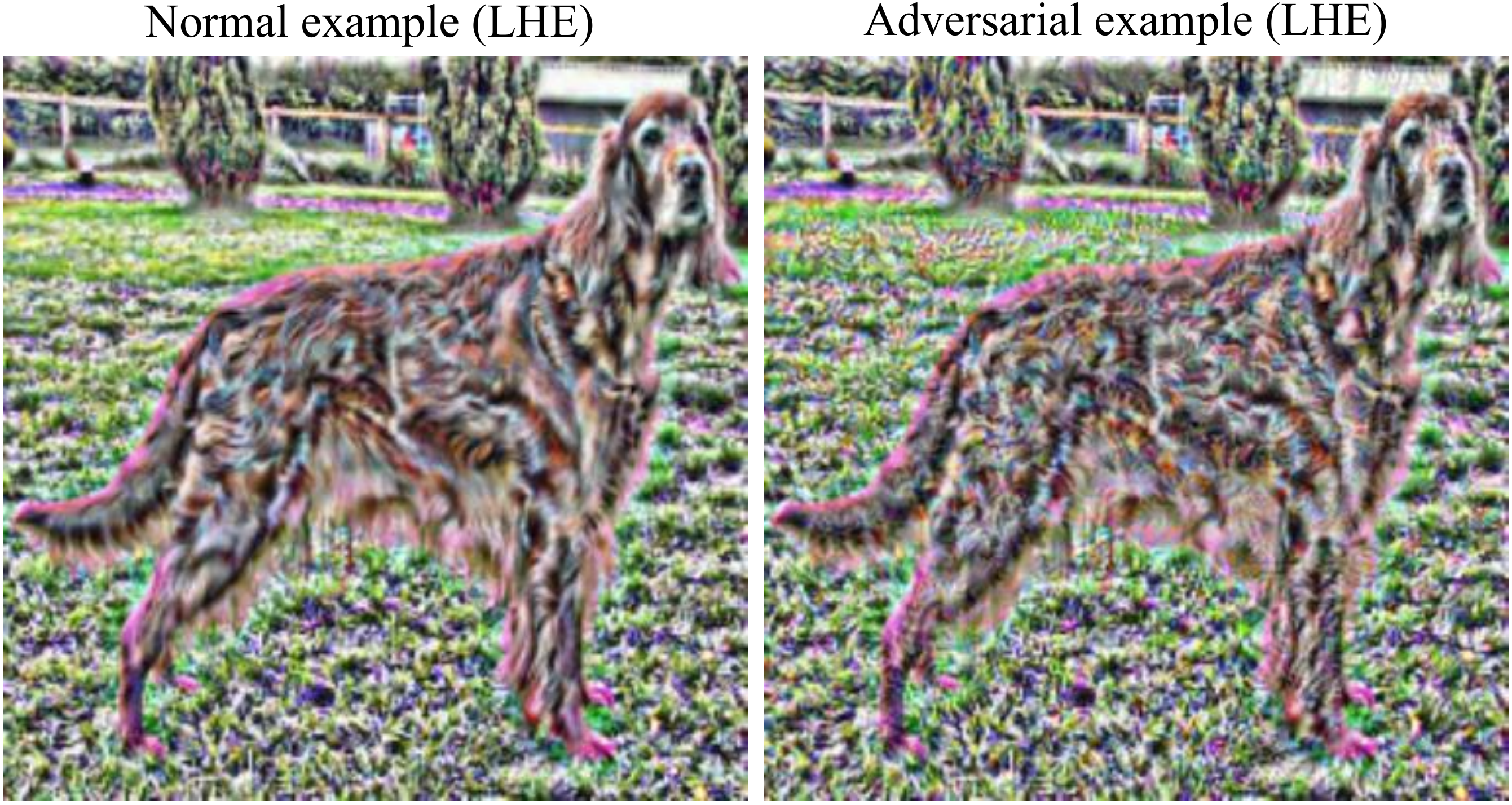} 
            \label{c}
        \end{minipage}
    }
    \subfigure[Feature-level with LHE]{
        \centering
        \begin{minipage}[b]{0.48\textwidth}
            \includegraphics[width=1\textwidth]{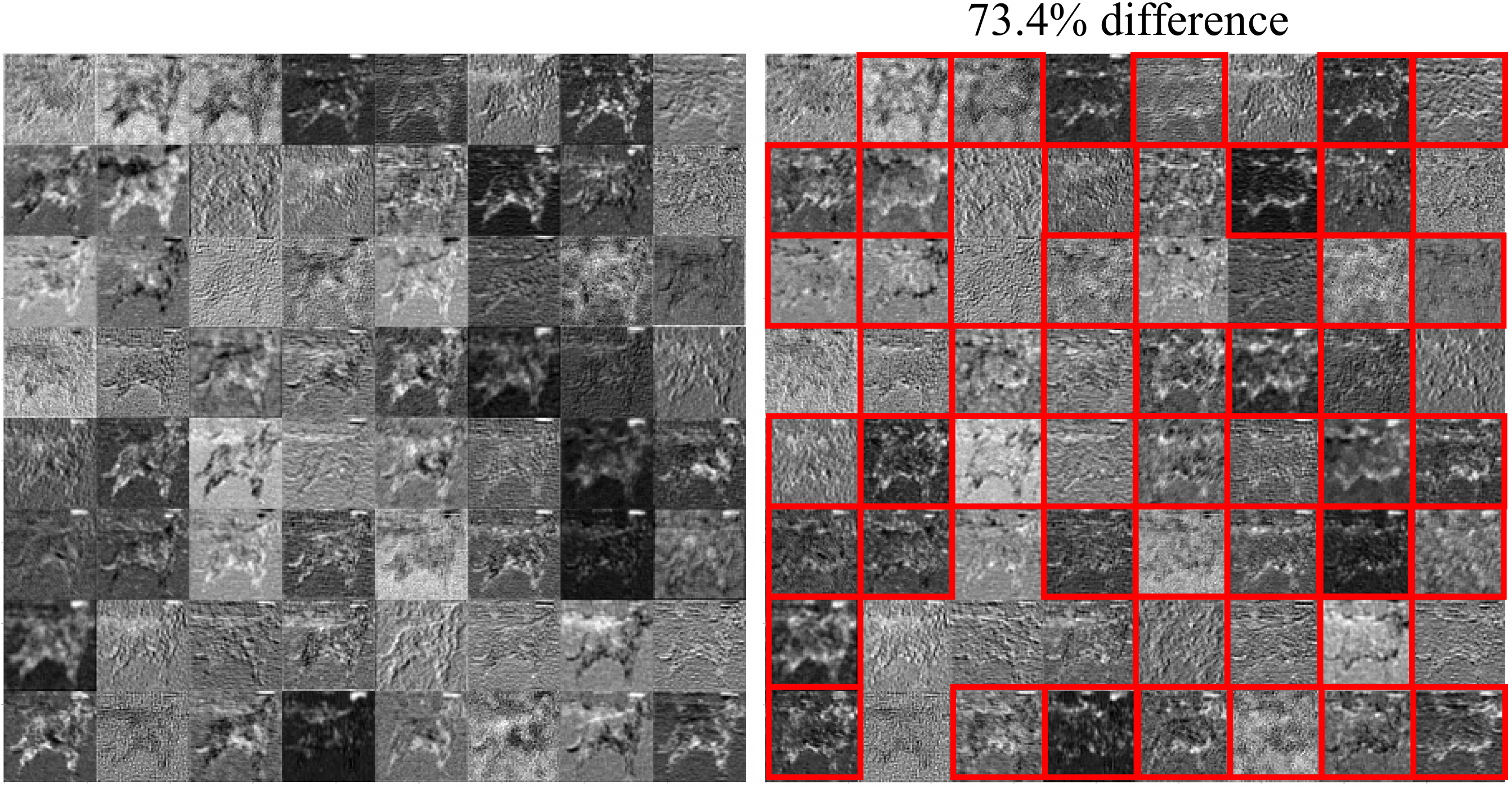}
            \label{d}
        \end{minipage}
    }
    \quad

    \caption{ Differences at pixel-level and feature-level between images with and without LHE.}
    \label{fig4}
\end{figure*}

\begin{figure}[!hb]
    \centering
    \includegraphics[width = 3.5 in]{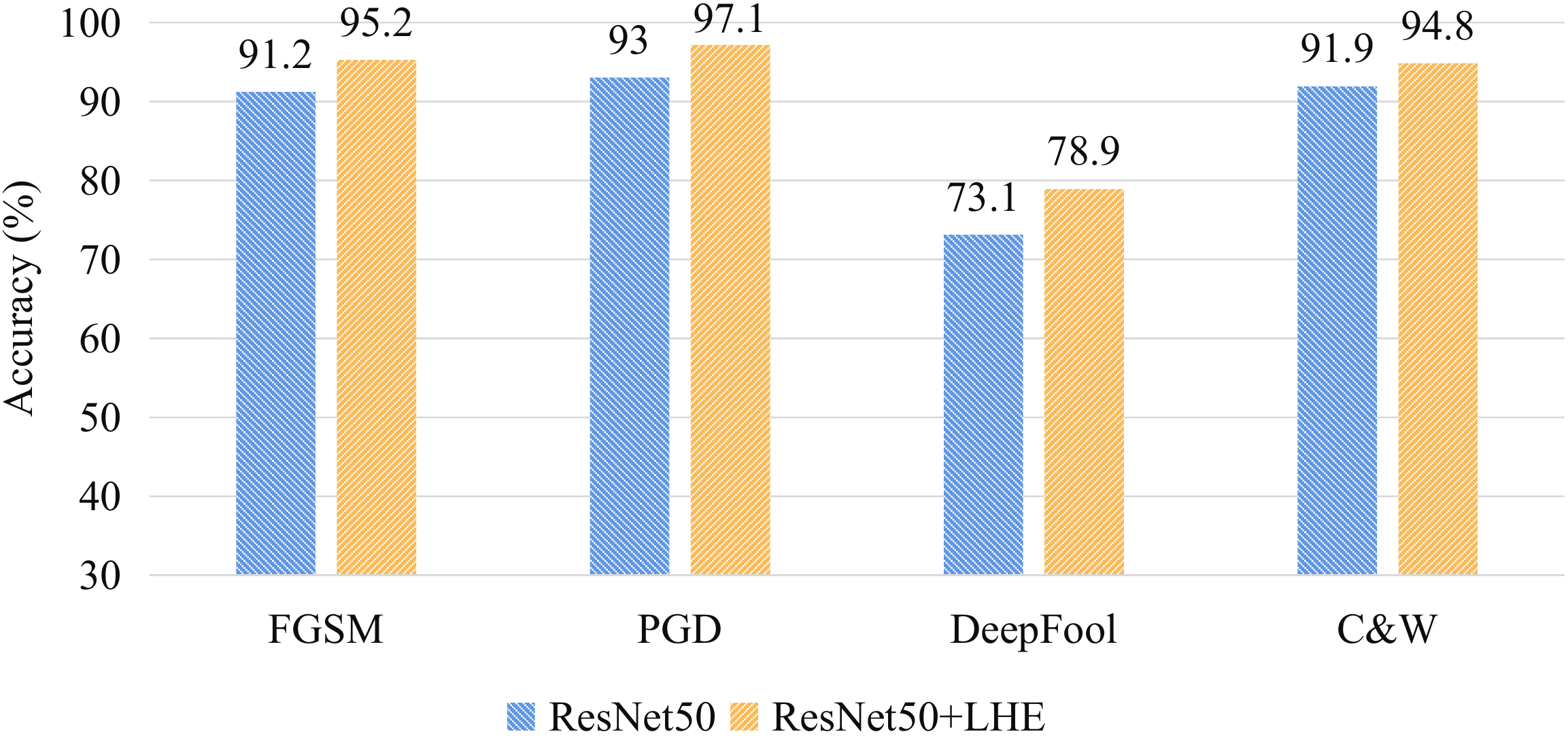}
    \caption{Performance comparison of detection training on ResNet50 on the dataset without LHE and the dataset with LHE.}
    \label{fig5}
\end{figure}

\begin{figure*}[!b]
    \centering
    \subfigure[High-Pass Filter-3D]{
        \centering
        \begin{minipage}[b]{0.29\textwidth}
            \includegraphics[width=1\textwidth]{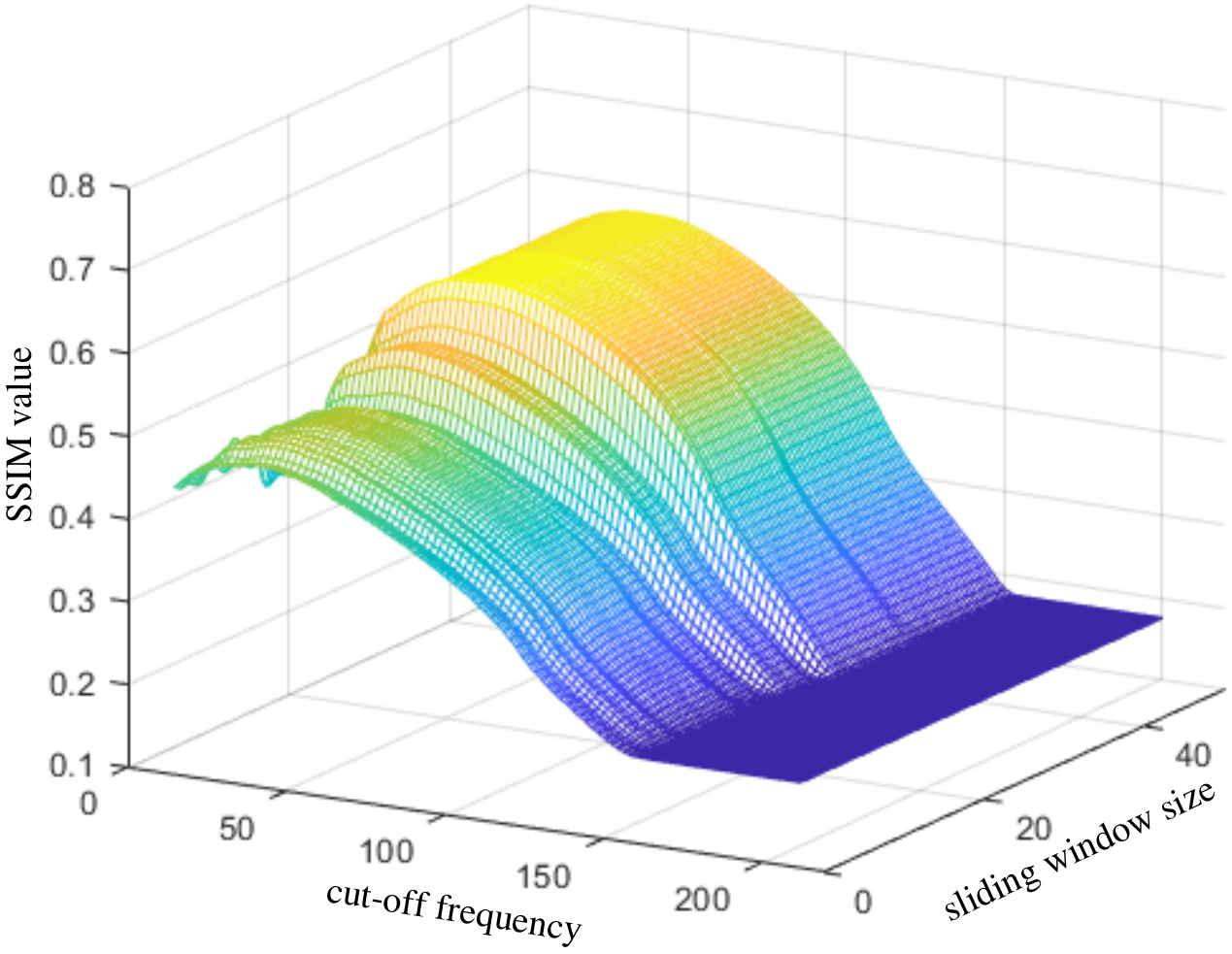} 
        \end{minipage}
    }
    \subfigure[Gaussian High-Pass Filter-3D]{
        \centering
        \begin{minipage}[b]{0.29\textwidth}
            \includegraphics[width=1\textwidth]{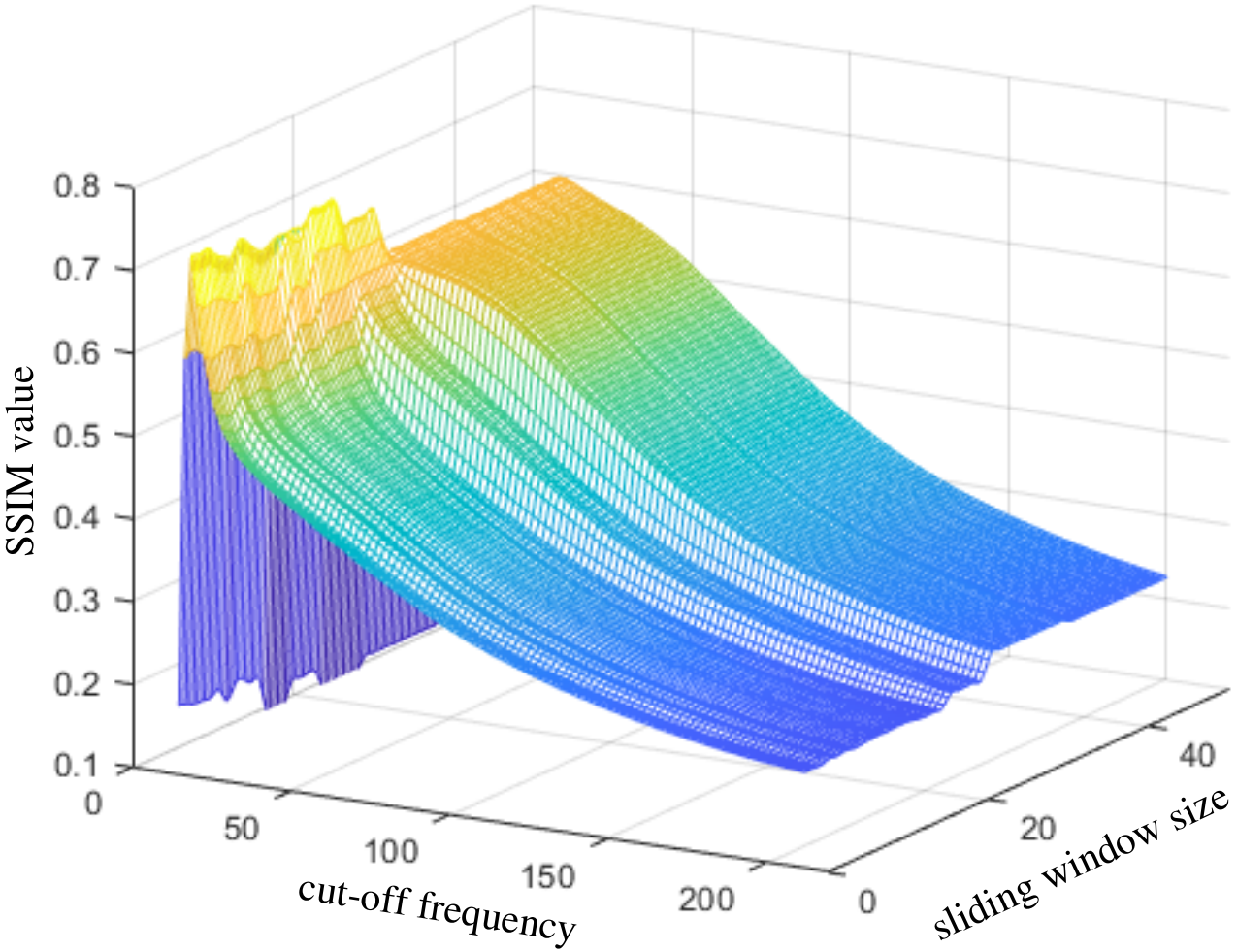} 
        \end{minipage}
    }
    \subfigure[Butterworth High-Pass Filter-3D]{
        \centering
        \begin{minipage}[b]{0.29\textwidth}
            \includegraphics[width=1\textwidth]{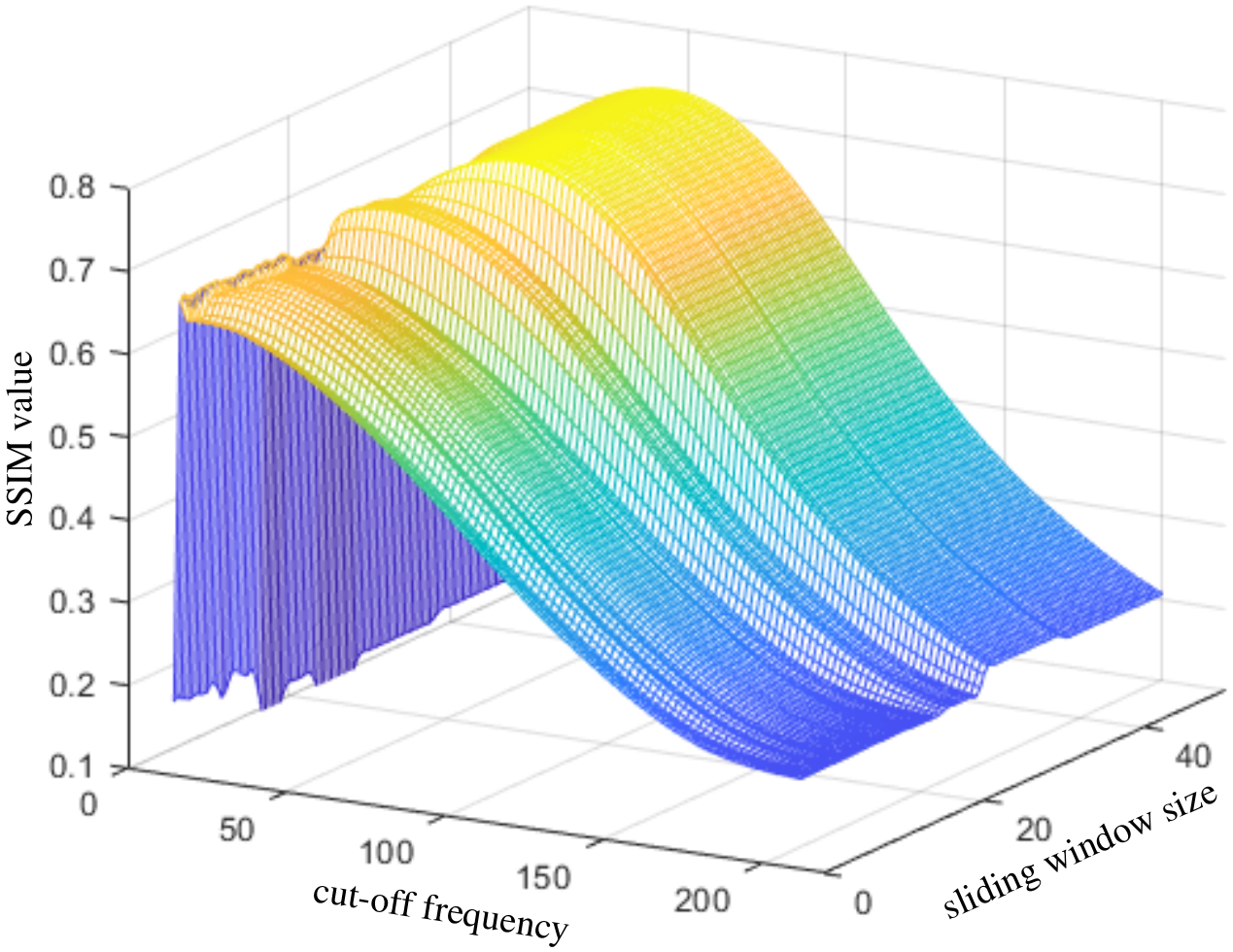} 
        \end{minipage}
    }
    \quad
    
    \centering
    \subfigure[High-Pass Filter-Section]{
        \centering
        \begin{minipage}[b]{0.29\textwidth}
            \includegraphics[width=1\textwidth]{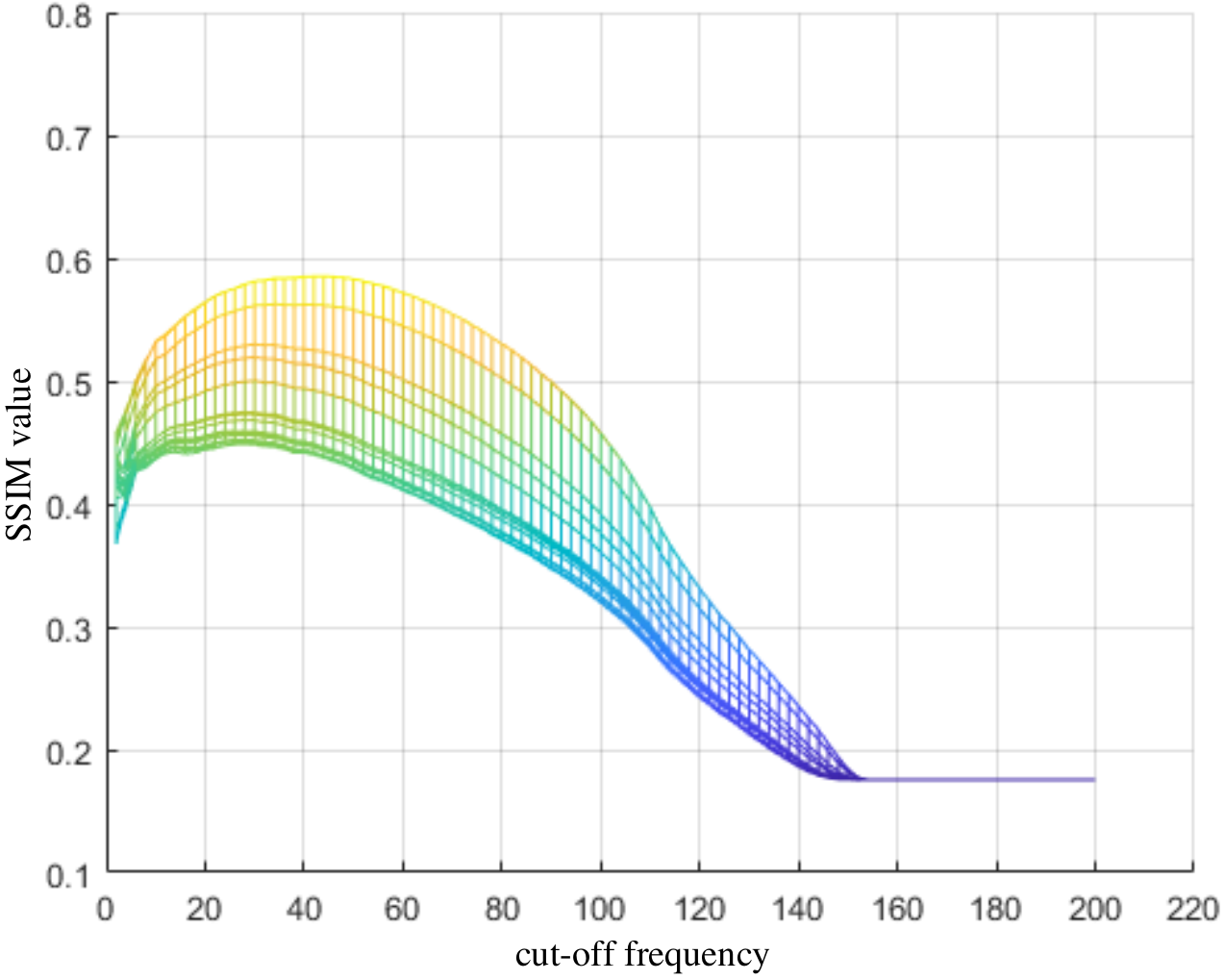} 
        \end{minipage}
    }
    \subfigure[Gaussian High-Pass Filter-Section]{
        \centering
        \begin{minipage}[b]{0.29\textwidth}
            \includegraphics[width=1\textwidth]{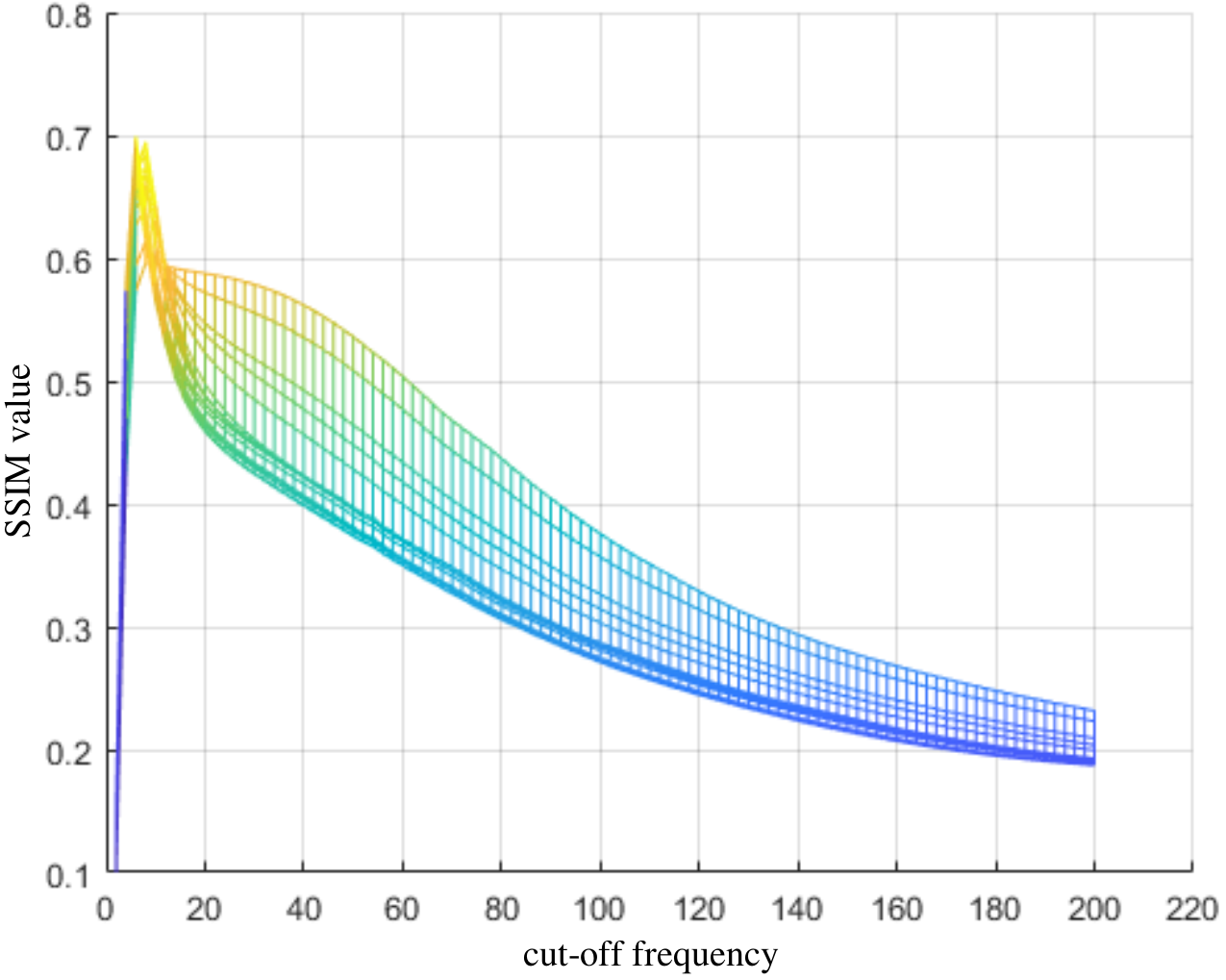} 
        \end{minipage}
    }
    \subfigure[Butterworth High-Pass Filter-Section]{
        \centering
        \begin{minipage}[b]{0.29\textwidth}
            \includegraphics[width=1\textwidth]{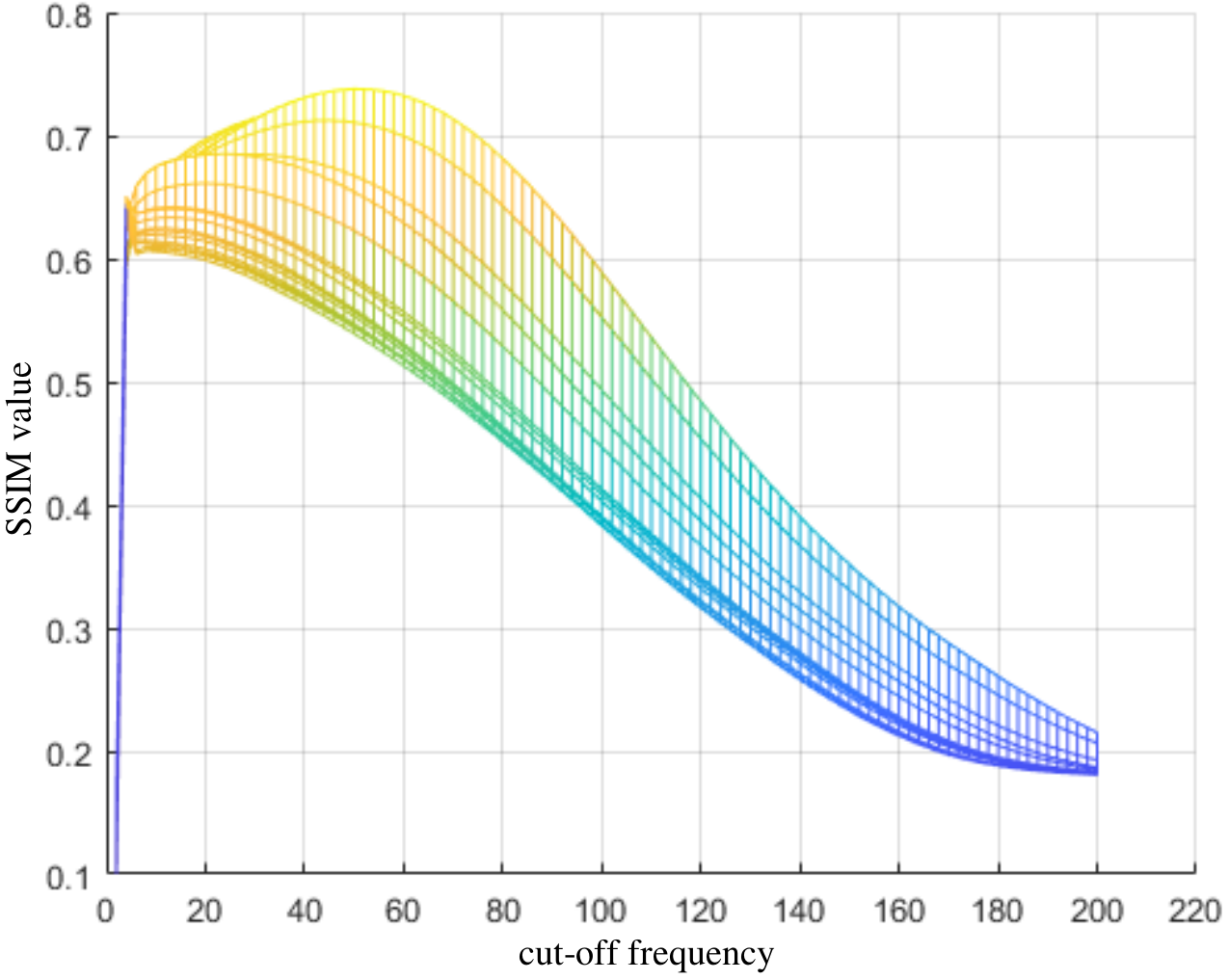} 
        \end{minipage}
    }
    \quad

    \centering
    \subfigure[High-Pass Filter-Top]{
        \centering
        \begin{minipage}[b]{0.29\textwidth}
            \includegraphics[width=1\textwidth]{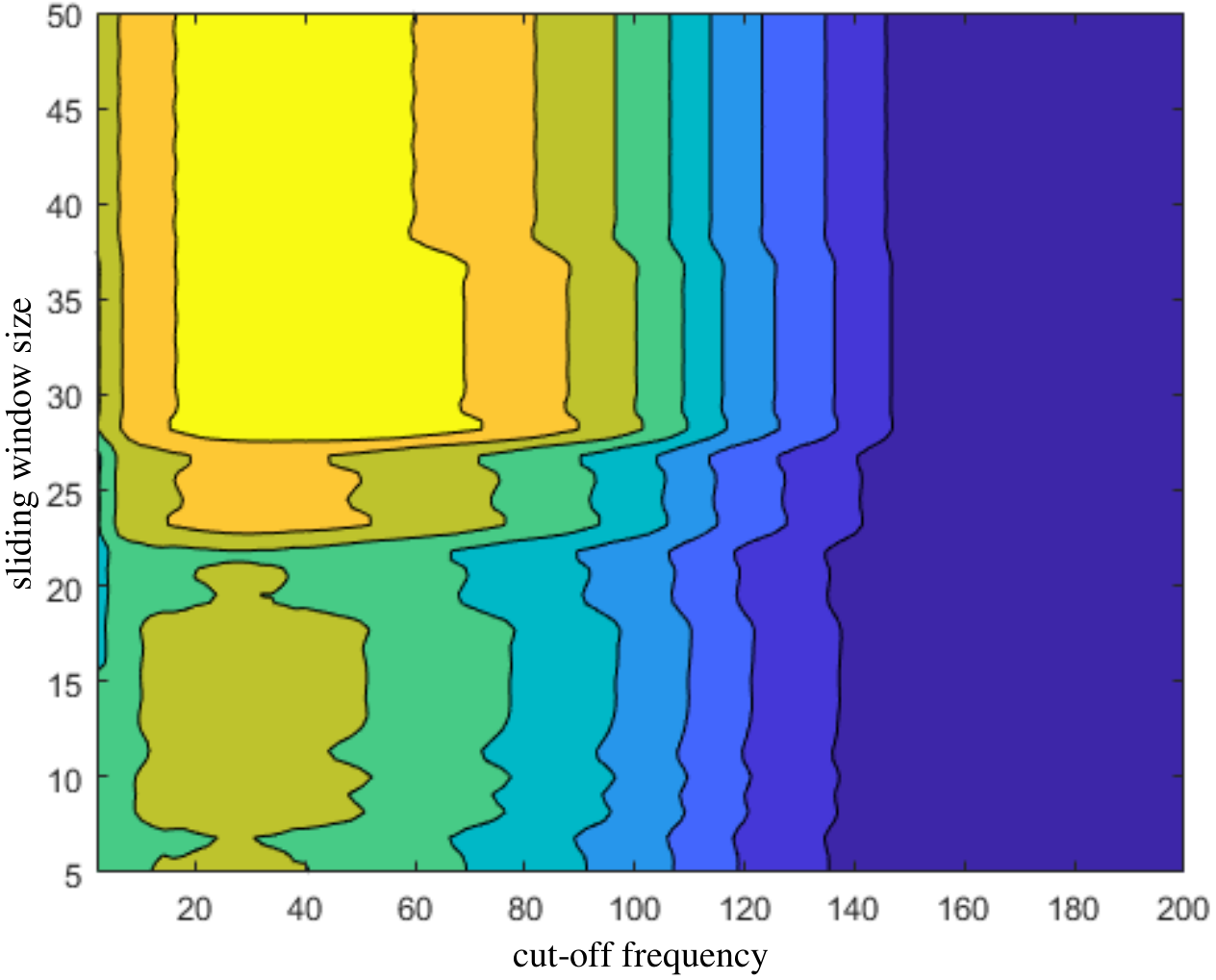} 
        \end{minipage}
    }
    \subfigure[Gaussian High-Pass Filter-Top]{
        \centering
        \begin{minipage}[b]{0.29\textwidth}
            \includegraphics[width=1\textwidth]{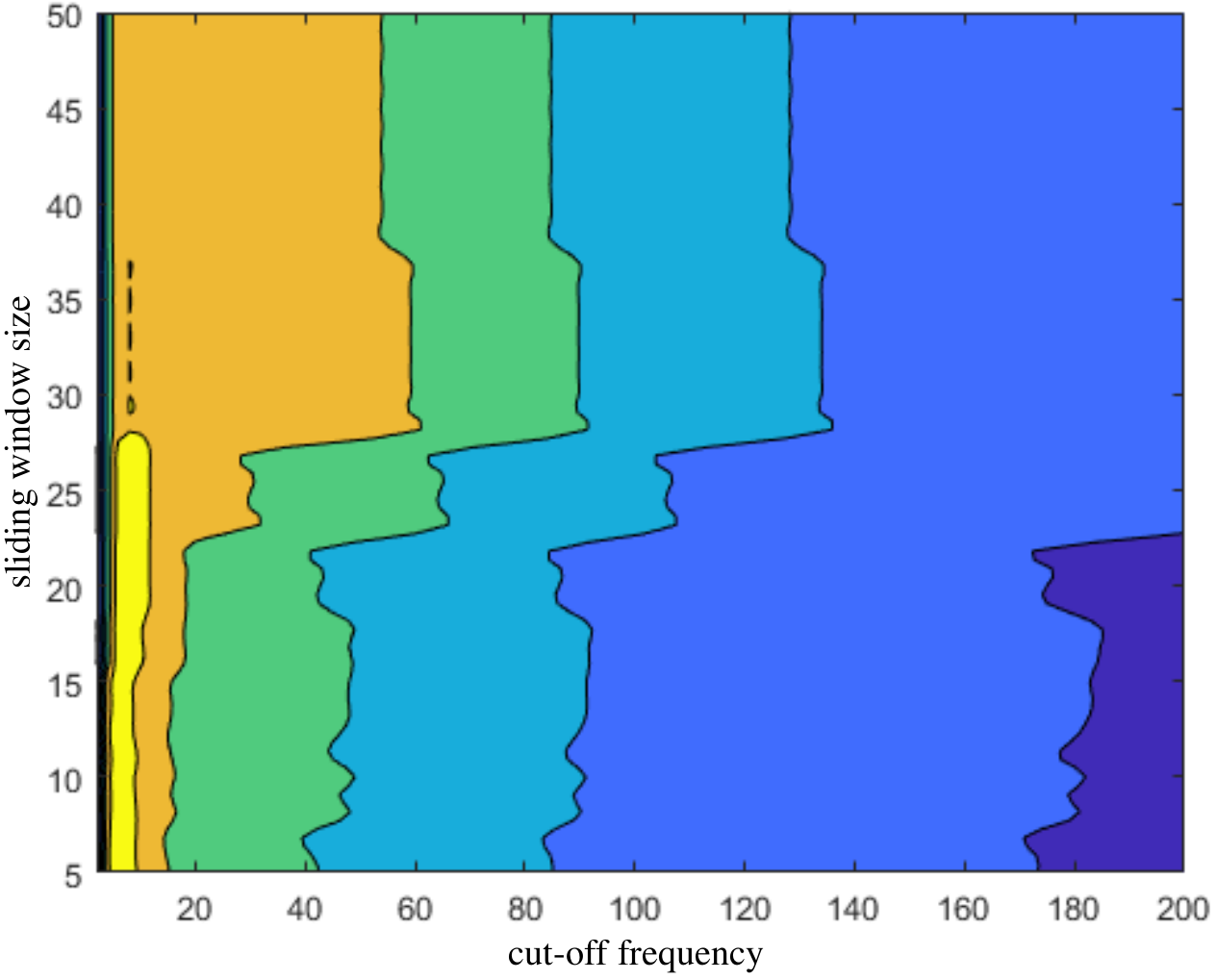} 
        \end{minipage}
    }
    \subfigure[Butterworth High-Pass Filter-Top]{
        \centering
        \begin{minipage}[b]{0.29\textwidth}
            \includegraphics[width=1\textwidth]{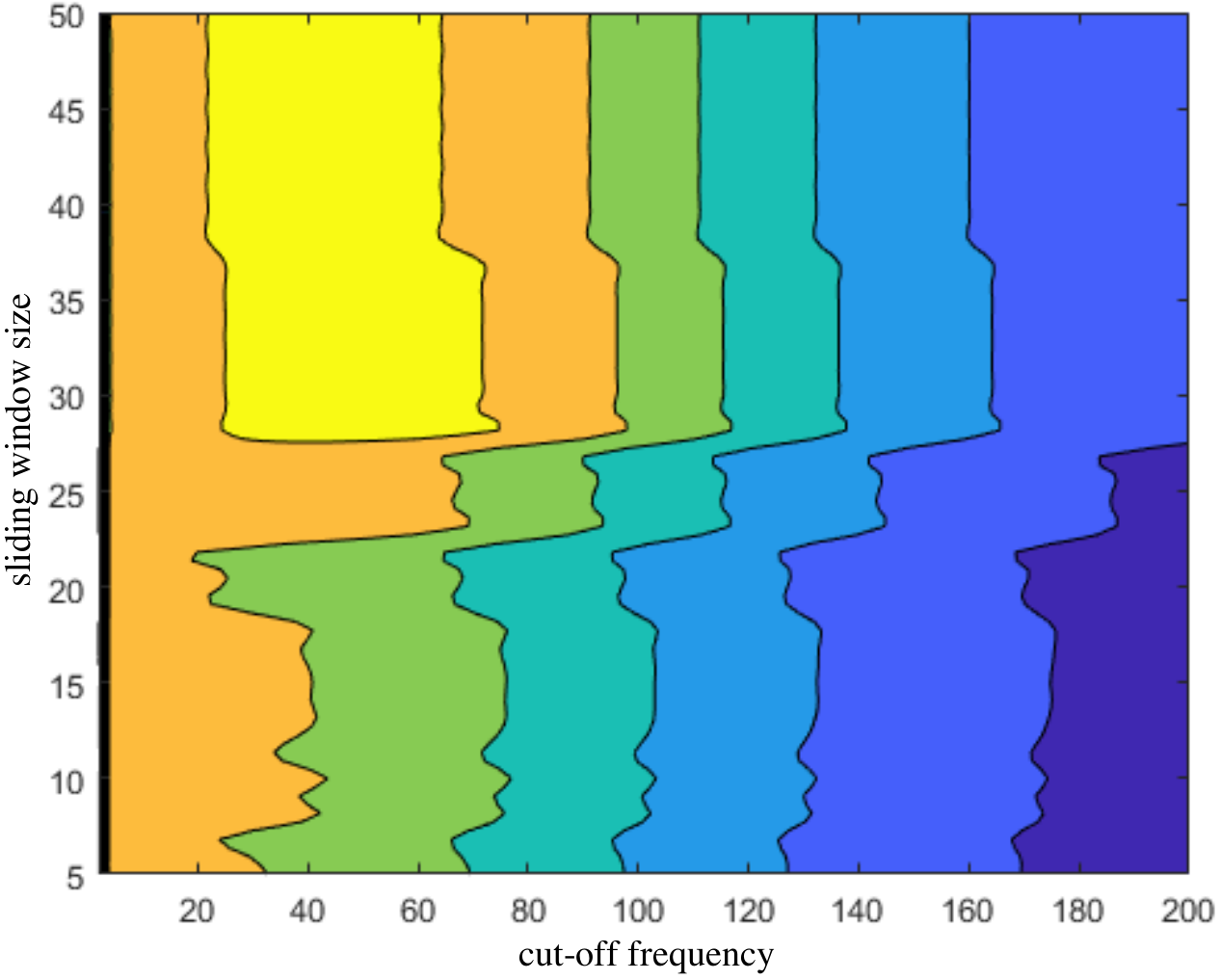} 
        \end{minipage}
    }
    \quad

    \caption{ (a)(b)(c) reflecting the effect of the filter cut-off frequency and  LHE sliding window size on SSIM value. (d)(e)(f) are cross-sectional diagrams reflecting the effect of the filter cut-off frequency on SSIM value. (g)(h)(i) are top views, which reflect the influence of filter cut-off frequency and LHE sliding window size on SSIM value. The highlighted part in the figure is the area with a considerable SSIM value. The experiment used images from part of the ImageNet dataset.}
    \label{fig6}
\end{figure*}

\subsection{Ablation study \label{sec:Ablation study}}

In this section, we first conduct ablation studies on the effects of different parameters of the proposed Image Contrast Enhancement module and High-frequency Information Extraction module. Secondly, the parameter selection and related settings are carried out. Finally, the effectiveness of the proposed method is analyzed. The section is divided into the following two parts.

\subsubsection{Ablation study on LHE \label{sec:Ablation study on LHE}}

\par As mentioned above, LHE considers the correlation of local pixels in an image more specifically. We sample a subset of LHE-processed normal and adversarial examples and compare them with the unprocessed original versions, both at the pixel level and the feature level of the model.

\par The experimental results are shown in Fig. \ref{fig4}, which shows the pixel-level and feature-level differences between examples without and with LHE processing. The fig. \ref{a} shows the pixel-level differences, which can be seen to be small and almost indistinguishable. Fig. \ref{c} is the feature-level difference. The feature difference map for examples without LHE is about 26.5\%. After using LHE, the feature difference map increased to 73.4\%. The magnification of the difference in feature maps can significantly improve the discrimination ability of the model. At the same time, we perform binary classification training on LHE-processed test data and unprocessed test data.

\par Fig. \ref{fig5} shows the performance comparison of CNN binary classification without LHE and LHE. It can be seen that, after using LHE, the detection performance of the binary classifier for different adversarial examples can be improved by about 4\%.

\par For normal examples, the LHE-processed image can increase the contrast of region-related pixels; specifically, the texture details are more apparent. For adversarial examples, since the adversarial perturbation itself is non-semantic and random noise, in the image processing by LHE, the regional pixels are cluttered, and it is not easy to distinguish the texture details. 

Although the LHE-processed image loses some contour information, it enhances texture details and makes abnormal pixels more obvious. Experimental results show that LHE on input examples can improve the model's ability to detect adversarial examples.

\subsubsection{Ablation study on Filter \label{sec:Ablation study on Filter}}

Although the LHE-processed adversarial examples have more abnormal pixels than normal examples, there are still too many interference factors for the classification model. We hope that the model can pay more attention to those pixels that are abnormal with the surrounding pixels rather than the whole image.

Given that adversarial perturbations are high-frequency information \cite{Duan_2021_ICCV}, it is a good idea to use a high-pass filter to filter out these adversarial perturbations and remove low-frequency information. The experiments in this section compare three commonly used filters and determine the optimal parameters during the experiment. The experiment randomly selects 500 images in the ImageNet \cite{krizhevsky2012imagenet} dataset and uses the PGD \cite{madry2018towards} attack method to generate their corresponding adversarial examples. Considering that different window sizes in the LHE processing process bring different results, the three experimental variables are the LHE window size, filter kind, and filter cut-off frequency. Since structural similarity (SSIM) \cite{wang2004image} can be used to measure the similarity between the extracted high-frequency information and the original adversarial perturbation, the result indicator is the numerical value of the structural similarity. The closer the value is to one, the higher the similarity.

\par The similarity measurement available from the SSIM measurement system can be composed of three contrast modules: luminance ($l$), contrast ($c$), and structure ($s$). The three functions are combined to get the SSIM exponential function
\begin{equation}
    SSIM(X, Y)=[l(X, Y)]^{\alpha}[c(X, Y)]^{\beta}[s(X, Y)]^{\gamma} ,
\end{equation}
among them, $X$ and $Y$ correspond to the original signal and the processed signal, $l(X, Y)$ compares the brightness of $X$ and $Y$, $c(X, Y)$ compares the contrast of $X$ and $Y$, and $s(X, Y)$ compares the structure of $X$ and $Y$, $\alpha>0$, $\beta>0$, $\gamma>0$ are parameters for adjusting brightness, contrast and structure.

\begin{figure*}[!htb]
    \centering
    \includegraphics[width = 6 in]{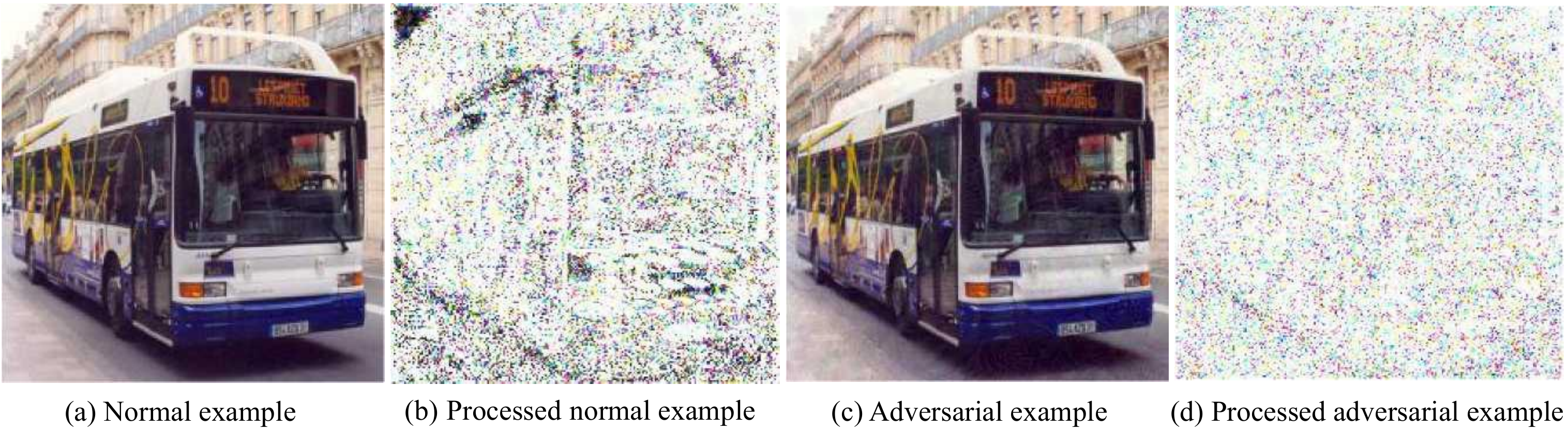}
    \caption{ Normal example, adversarial example, and examples after extracting high-frequency information (20x magnification).}
    \label{fig7}
\end{figure*}

As seen from the first row of Fig. \ref{fig6}, when the cut-off frequency of the high-pass filter increases from 0 to 100, the curve is relatively flat; from 100 to 150, the curve gradually becomes steeper, and there is no change after 150. The Gaussian high-pass filter is very steep near the cut-off frequency of 0 to 5, rapidly decreases from 5 to 10, and starts to flatten at 10. The too-steep curve is not conducive to the generalization of the filter to different images. The Butterworth high-pass filter has a smooth descending curve that can reach the highest SSIM value among the three. As the size of the sliding window increases, the variation of the Butterworth high-pass filter remains stable, which means that choosing a specific cut-off frequency has better extraction performance for different images. Therefore, we use the Butterworth high-pass filter as the high-frequency information extractor in the experiments. It can be seen from the figure that the cut-off frequency is 50-60, and the sliding window size of 25-30 can be used for better setting parameters.

The result after LHE and high-frequency information extraction is shown in Fig. \ref{fig7}. For normal examples, the extracted high-frequency information includes the contour information of the object. In contrast, the high-frequency information extracted from adversarial examples is semantically free and randomly distributed. The difference in the processed features is significant, which explains why the method can improve the model's detection ability and proves the method's effectiveness.

\section{Conclusion}

\par We experimentally observe that adversarial perturbations differ in local correlation from high-frequency information in normal examples. This paper proposes an adversarial examples detection method that exploits this correlation difference and aids in detection by enhancing image contrast. The contrast of high-frequency information is first enhanced by local histogram equalization. High-frequency features are extracted using a Butterworth high-pass filter to improve classifier performance. The experimental results show that the local histogram equalization method using local correlation can effectively enhance the difference between adversarial examples and normal examples. At the same time, the method can accurately detect common adversarial attacks. It not only has significant advantages in detecting UAP \cite{moosavi2017universal} attacks, but also performs well in cross-models attack detection.   
\par This work enhances the feature differences between adversarial and normal examples through image contrast enhancement, which can be considered for future perturbation removal studies.

\ifCLASSOPTIONcompsoc
  \section*{Acknowledgments}
\par This research work is partly supported by National Natural Science Foundation of China No.62172001 and No.62076147.

\ifCLASSOPTIONcaptionsoff
  \newpage
\fi

\bibliographystyle{IEEEtran}
\bibliography{mybib}

\par\noindent 
\parbox[t]{\linewidth}{
\noindent\parpic{\includegraphics[height=1.5in,width=1in,clip,keepaspectratio]{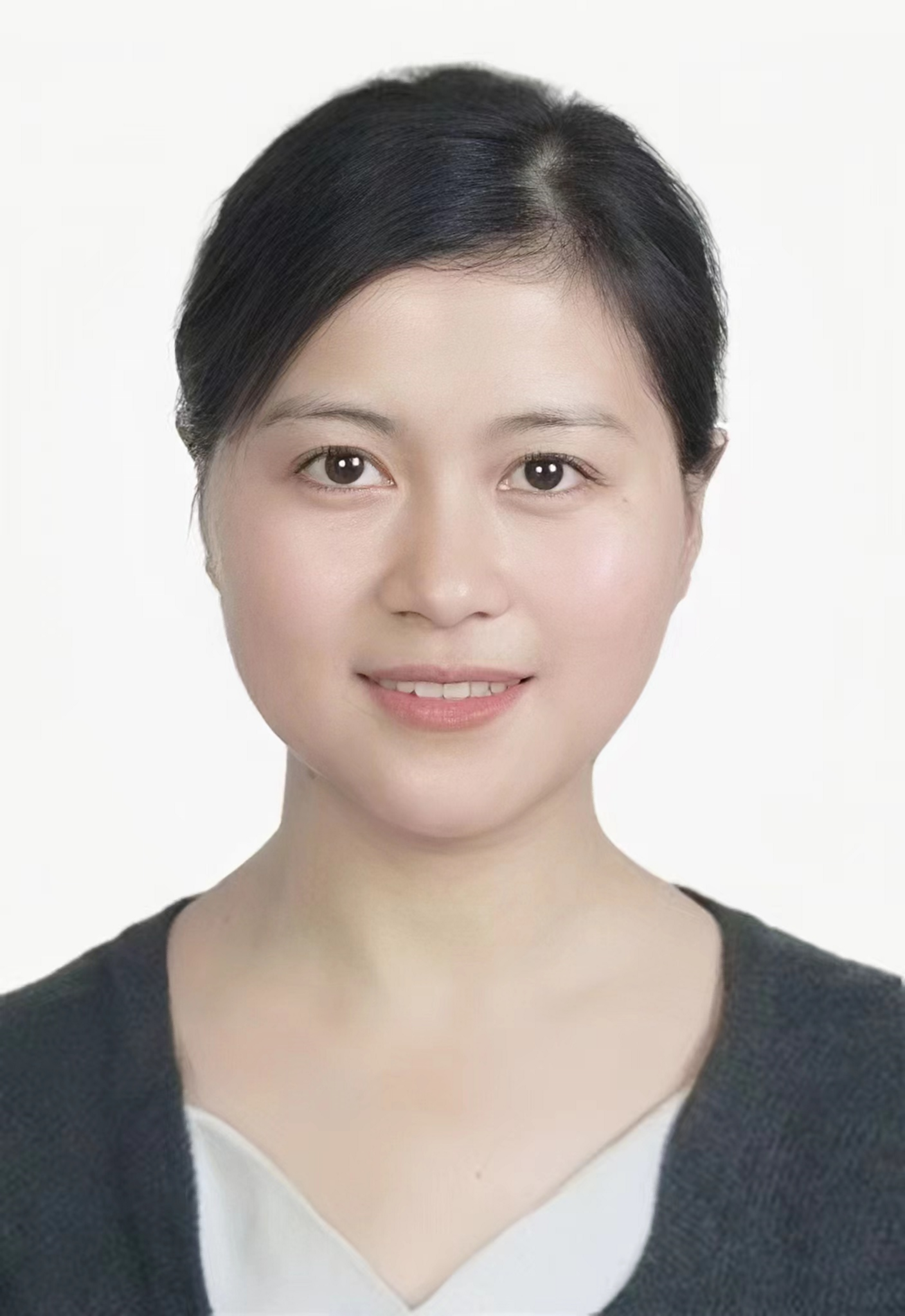}}
\noindent {\bf Zhaoxia Yin}
received her B.Sc., M.E. \& Ph.D. from Anhui University in 2005, 2010 and 2014 respectively. She was an Associate Professor and Doctoral Tutor in the School of Computer Science and Technology at Anhui University. She was also with Purdue University as a Visiting Scholar from 2017 to 2018. Currently she works as a full professor in the School of Communication \& Electronic Engineering at East China Normal University. Her research interests include multimedia \& AI security, image processing, and digital forensics. She has published over 100 research papers \& patents and is the Principal Investigator of three NSFC research Projects. Email: zxyin@cee.ecnu.edu.cn.}
\vspace{0\baselineskip}

\par\noindent 
\parbox[t]{\linewidth}{
\noindent\parpic{\includegraphics[height=1.5in,width=1in,clip,keepaspectratio]{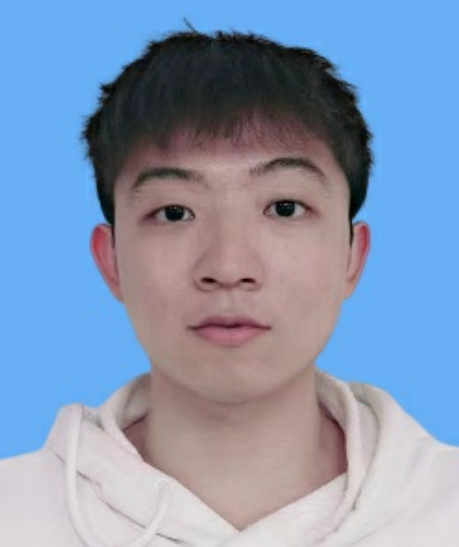}}
\noindent {\bf Shaowei Zhu}
received his bachelor degree in computer science and technology in 2019 and is currently a master student at the School of Computer Science and Technology, Anhui University. His current research interests include AI security and adversarial example defense research.
Email: zhusw520@gmail.com.}
\vspace{0\baselineskip}

\par\noindent 
\parbox[t]{\linewidth}{
\noindent\parpic{\includegraphics[height=1.5in,width=1in,clip,keepaspectratio]{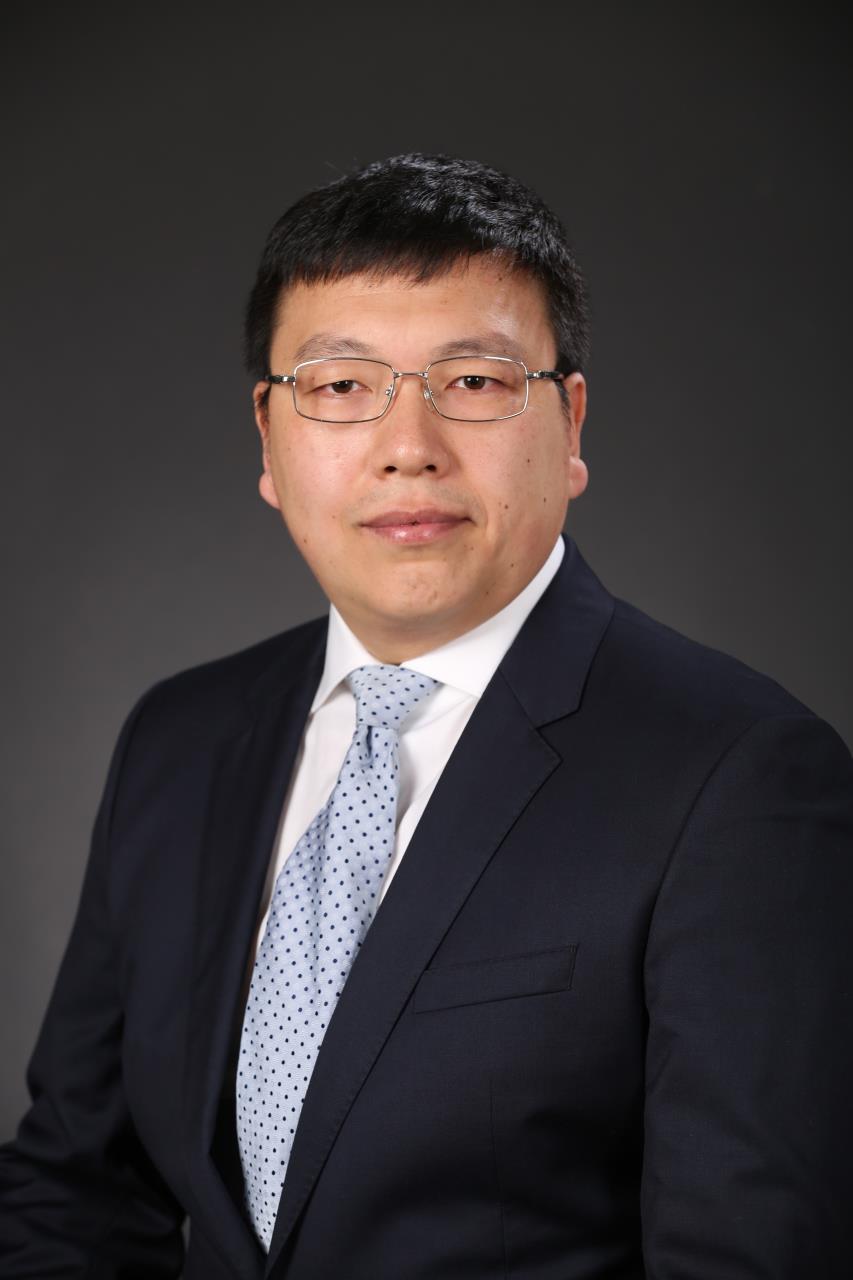}}
\noindent {\bf Hang Su}
IEEE member, is an associated professor in the department of computer science and technology at Tsinghua University. His research interests lie in the development of computer vision and machine learning algorithms for solving scientific and engineering problems arising from artificial learning and reasoning. His current work involves both the foundations of adversarial machine learning, based on which he has published around 50 papers including CVPR, ECCV, TMI, etc. He has served as area chair in NeurIPS and the workshop co-chair in AAAI22. He received “young investigator award” from MICCAI2012, the “best paper award” in AVSS2012, and “platinum best paper award” in ICME2018. }
\vspace{0\baselineskip}

\par\noindent 
\parbox[t]{\linewidth}{
\noindent\parpic{\includegraphics[height=1.5in,width=1in,clip,keepaspectratio]{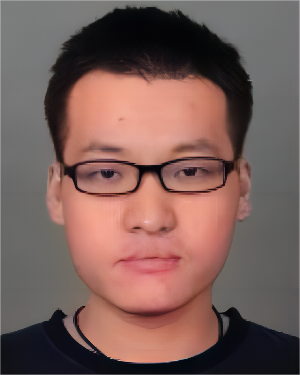}}
\noindent {\bf Jianteng Peng}
is currently a senior computer vision algorithm engineer in OPPO Intellisense and Interaction Research Department. He received his doctorate from the School of Computer Science, Beijing Institute of Technology. Peng Jianteng has authored about CV journals and conference papers in international journals and conferences, such as CVPR, TNN, TIP, CSVT, TCYB, etc. Now, Peng Jianteng is interested in face recognition, face cluster, face generation, etc. The face algorithms developed by his team are applied to the album clustering in OPPO phones. Email: pengjianteng@oppo.com.}
\vspace{2\baselineskip}

\par\noindent 
\parbox[t]{\linewidth}{
\noindent\parpic{\includegraphics[height=1.5in,width=1in,clip,keepaspectratio]{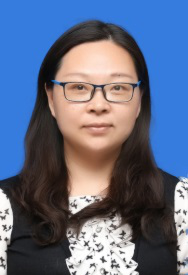}}
\noindent {\bf Wanli Lyu}
received the MS degree in computer science and technology with Guangxi University and the PhD degree in computer science and technology with Anhui University. She was a postdoctoral research fellow in Department of Information Engineering and Computer Science at Feng Chia University from August 2013 to July 2014. Since July 2004, she is a Lecturer in School of Computer Science and Technology, Anhui University. Her current research interests include image processing, data hiding and information security.}
\vspace{2\baselineskip}

\par\noindent 
\parbox[t]{\linewidth}{
\noindent\parpic{\includegraphics[height=1.5in,width=1in,clip,keepaspectratio]{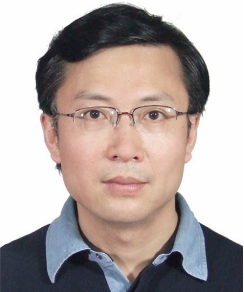}}
\noindent {\bf Bin Luo}
received his BEng. and MEng. degrees in electronics from Anhui university, China. In 2002, he was awarded the PhD degree in Computer Science from the University of York, UK. He is currently a full professor at Anhui University. He is the chair of IEEE Hefei Subsection, and an associate chair of IAPR TC15. He serves as the editor-in-chief of the Journal of Anhui University (Natural Science Edition), an associate editor of several international journals, including Pattern Recognition, Pattern Recognition Letters, Cognitive Computation and International Journal of Automation and Computing. He was the guest editors for the Journal Special Issue of the Pattern Recognition Letters and Cognitive Computation. His current research interests include pattern recognition and digital image processing. In particular, he is interested in structural pattern recognition, graph spectral analysis, image and graph matching. He has published about 500 research papers in journals, edited books and refereed conferences. Some of his papers were published in the journals of IEEE TPAMI, IEEE TIP, Pattern Recognition, Pattern Recognition Letters and Neurocomputing, and the conferences of CVPR, NIPS, IJCAI and AAAI conferences.}
\vspace{2\baselineskip}

\end{document}